\documentclass[11pt]{article}
\usepackage[T1]{fontenc}
\usepackage{lmodern}
\usepackage{natbib}
\setcitestyle{authoryear,round,citesep={;},aysep={,},yysep={;}}

\usepackage{amssymb}
\usepackage{amsmath}
\usepackage{amsthm}
\usepackage{graphicx}
\usepackage{hyperref}
\usepackage{url}
\usepackage{array}
\usepackage{tabularx}
\usepackage[margin=1in]{geometry}
\newcolumntype{Y}{>{\raggedright\arraybackslash}X}

\theoremstyle{plain}
\newtheorem{theorem}{Theorem}
\newtheorem{proposition}{Proposition}
\newtheorem{lemma}{Lemma}
\newtheorem{corollary}{Corollary}

\title{Minimal Superior Penalties: Characterising and Calibrating Decision-Aware Proper Scoring Rules}
\author{Rouhollah Ahmadian \\
  Department of Mathematics and Computer Science \\
  Amirkabir University of Technology (Tehran Polytechnic), Iran \\
  \texttt{rahmadian@aut.ac.ir}}
\date{}

\begin{document}

\maketitle

\begin{abstract}
Training a classifier means letting a validation metric decide which model is kept and when training stops. A natural requirement on that metric is superiority: every correctly classified forecast should score better than every incorrectly classified one, however close the underlying probabilities are. Neither the Brier score nor the logarithmic score satisfies this on its own. We show that superiority is governed by the separation between two endpoints -- the supremum of the score on correct predictions and the infimum on incorrect predictions -- and derive both in closed form for each score. Their difference is the infimal penalty, the greatest lower bound on the penalty needed to restore superiority. Prior work restores superiority with a penalty calibrated from the first endpoint alone; for the Brier base, our infimal penalty is smaller than that published penalty by a factor ranging from 4.00 at three classes to 2.11 at twenty. We further embed the published and infimal penalties in a one-parameter family and show that, under a single-valued tie convention, every member of the family is strictly proper. In a pre-specified two-dataset case study, mean selected-model macro-F1 rises with the penalty in the reported ten-class regimes, though not distinguishably from noise once seed/selection-mode nesting is accounted for; a fifth regime at forty-seven classes does not reproduce even this descriptive pattern. The intended no-superiority baseline is algebraically a rescaling of the family under test, so the experiment cannot attribute the observed pattern specifically to superiority. When superiority itself is the objective, any penalty strictly above the infimal value suffices; choosing a larger penalty is a separate calibration trade-off rather than a requirement for the guarantee.

\end{abstract}


\section{Introduction}
\label{sec:introduction}

Training a classifier means watching a validation metric decide, epoch by
epoch, which checkpoint is kept and when training stops — and that metric,
not any criterion the practitioner separately examined, is what determines
which model ships. A natural requirement on such a metric — though not by
itself a guarantee that it favours the better checkpoint on average — is a
kind of fairness toward the decision it is meant to inform: a scoring rule
used this way should never rank a wrong classification ahead of a right
one, however close the underlying probabilities are. Stated precisely, the
requirement is that the rule's range of achievable scores over correctly
classified forecasts must not overlap its range over incorrectly classified
ones. It constrains the score on each individual prediction, not the
average score a validation set produces or which checkpoint that average
selects. This paper is about that pointwise requirement, when it holds,
and what it costs when it is enforced more heavily than it needs to be. We
treat it as one guarantee worth having among others, not as a sufficient
condition for good model selection and not as the sole explanation for
whatever macro-F1 benefit a penalised metric happens to show in practice —
our own experiment sets out to test that distinction directly, and what it
finds, including where the test itself falls short, is part of what this
paper reports.

\citep{ahmadian2025superior} identify exactly this
failure in the Brier score, the workhorse of probabilistic multi-class
evaluation. In their own worked example, a correctly classified forecast
scores $0.653$ under sum-form Brier, while an incorrectly classified one —
the vector $[0.51,0.49,0]$, a near-tie that just barely favours the wrong
class — scores $0.520$: lower, hence better by the raw metric, despite being
wrong. They name the property that rules this out \textit{superiority}: a scoring
rule is superior if its score on every correct prediction beats its score on
every incorrect one — a pointwise guarantee about individual predictions,
not a claim about the average score a checkpoint gets on a validation set
or about which checkpoint that average favours — and they show that the
Brier score and the logarithmic score are not superior on their own, so
each needs a penalty added on misclassification to become so. Their
observation is correct, and their fix restores exactly the ordering a
validation metric needs at the level of individual predictions. What is
left open is not whether to add a penalty, but how large one has to be.

\begin{figure}[htbp]
\centering
\includegraphics[width=3.2in]{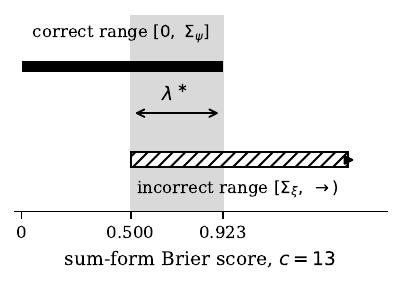}
\caption{The range of sum-form Brier scores attainable on correctly
classified forecasts, $[0, \Sigma_\psi]$, and on incorrectly classified
ones, $[\Sigma_\xi, \to)$, at $c=13$. The two ranges overlap (shaded); the
overlap has width $\lambda^\ast=\Sigma_\psi-\Sigma_\xi=0.423$, the infimal
penalty that separates them. $\Sigma_\psi=0.923$ and $\Sigma_\xi=0.500$
follow from the closed forms established in Section~\ref{sec:theory-1}; the
hatched bar is drawn open-ended since the incorrect range is unbounded
above.}
\label{fig:overlap}
\end{figure}

Superiority is a condition on two numbers: the largest score a rule can
achieve on a correct prediction, and the smallest it can achieve on an
incorrect one (Figure~\ref{fig:overlap}). The source paper's penalty is calibrated from the first number
alone — the correct-prediction range's own maximum — with the second left
implicit at zero. This conflates two effects that are not the same: enforcing
the ordering superiority requires, and re-weighting the metric toward raw
accuracy. A penalty large enough to do the first is not thereby the smallest
one that does it, and every unit of penalty beyond the smallest is
calibration sensitivity spent for no further purpose — the metric moves
further from measuring how well-calibrated a forecast is and closer to simply
counting errors. The second number, the smallest achievable score on an
incorrect prediction, turns out to have a closed form of its own, one the
source paper does not compute — and the configuration that attains it is the
limiting form of the source paper's own motivating vector above: its
intuition pointed directly at the quantity its analysis did not compute.
Subtracting the second number from the first yields a strictly smaller
penalty that still guarantees superiority. The gap is not small: the
published penalty exceeds this infimal value by a factor that runs from
$4.00$ at three classes down to $2.11$ at twenty — and the released
implementation, which pairs the same published penalty value with a
mean-form rather than sum-form base rule, exceeds it further still, by a
factor in the tens to hundreds depending on class count (relation to
published rules).

We treat superiority as an interval-separation condition between the two
ranges above, rather than as a property checked case by case, and compute
both endpoints in closed form for the Brier and logarithmic bases. This
yields the infimal penalty directly, as the gap between the two endpoints,
and recovers — as a consequence of the closed forms rather than a separately
argued case — the source paper's own restriction to more than two classes.
It also places the source paper's penalty as one member of a one-parameter
family of penalised rules — the specific finite penalty obtained by
calibrating from the first number alone, strictly above the infimum
$\lambda^\ast$ but with no admissible upper bound to be the largest of —
rather than as a fixed design choice. Every member of this family with a
large enough penalty is
strictly proper, for either base; which penalty values qualify depends on how
the family resolves an exact tie in the forecast, a dependence we work out
explicitly rather than leave implicit. We pre-specify an experimental
design to measure what the infimal penalty costs or saves in practice,
against the source paper's own choice, and intended a further baseline to
isolate error-oriented re-weighting from the ordering property itself; on
inspection that baseline turned out to already carry the ordering property
for every weight that drove its own observed rise, so this design does not
in fact isolate the two (Section~\ref{sub:results-baselines-does-the-frontier-require-superiority-or-only-re-weighting}; Section~\ref{sec:discussion}). The result we
can report is a negative one for the framing above, though we treat it as
descriptive rather than statistically confirmed once seed/mode nesting is
accounted for (Section~\ref{sub:results-primary-comparison-supercritical-vs-published-macro-f1-cell-level-analysis}): mean selected-model macro-F1
rises across the whole grid, all the way to the published value and
beyond, rather than flattening once $\lambda$ clears $\lambda^\ast$. This
observed pattern holds across four regimes at ten classes; a fifth, at
forty-seven classes, does not reproduce it, and we report both rather
than let the larger, more data-rich result stand in for a class-count
question this design was not
built to answer alone.

This paper's contributions, each delivered by one section, fall in a
deliberate order of priority. The first three are self-contained
theoretical results that hold on their own terms, independently of the
source paper; the fourth positions that theory against the source paper's
own propriety argument; the fifth is a pre-specified empirical case study
testing what the theory implies in practice, offered as a bounded
illustration alongside the theory rather than as a further theoretical
claim in its own right.

\begin{itemize}
\item A characterisation of superiority as separation between two closed-form ranges, for both the Brier and logarithmic scores.
\item The infimal penalty that induces superiority for each base, in closed form, together with the published penalty's exact position relative to it.
\item A one-parameter family of penalised rules that is strictly proper throughout, with the role of tie-handling in that guarantee made explicit.
\item A corrected argument for the source paper's own propriety theorem, holding for the whole family rather than the two published instances.
\item A pre-specified experiment measuring the published penalty's cost against the infimal one: at ten classes, mean selected-model macro-F1 rises well past the infimal penalty in the reported regimes, all the way to the published value and beyond, though this pattern is not statistically distinguishable from noise at the cell level once seed/mode nesting is accounted for — and the intended no-superiority baseline for isolating why turned out, on inspection, to already carry the ordering property itself, so this design does not establish the mechanism either — while a fifth regime at forty-seven classes does not reproduce the pattern.
\end{itemize}

Section~\ref{sec:preliminaries} fixes notation and the tie convention;
Section~\ref{sec:related-work} positions the contribution among proper
scoring rules, decision-theoretic evaluation, and the model-selection
criteria this paper's own penalty is one instance of, before the technical
sections that follow; Section~\ref{sec:theory-1} develops the
characterisation and the endpoints; Section~\ref{sec:theory-2} introduces
the family and establishes its propriety;
Section~\ref{sec:relation-to-published} positions the published rules
within it and repairs the propriety proof;
Section~\ref{sec:experimental-design} pre-registers the empirical test;
Section~\ref{sec:results} reports its outcome; Section~\ref{sec:discussion}
sets out what that outcome does and does not establish, together with
directions the family suggests but does not itself resolve;
Section~\ref{sec:limitations} states what bounds all of the above, without
exception for findings that cost us; Section~\ref{sec:conclusion} closes
the paper.


\section{Preliminaries}
\label{sec:preliminaries}

We consider single-label multi-class classification with $c\ge2$ classes. A
scoring rule assigns to a forecast $p$ and a realised class $\omega$ a real
number $S(p,\omega)$; we take every scoring rule in this paper to be
negatively oriented, so that smaller values indicate a better forecast.
We write $Q$ for a candidate true distribution over classes and
$S(P,Q):=\mathbb{E}_Q[S(P,\cdot)]$ for the expected score of a forecaster who
reports $P$ when the true distribution is $Q$. A scoring rule is \textit{proper} if
$S(Q,Q)\le S(P,Q)$ for every $P,Q$, and \textit{strictly proper} if the inequality is
strict whenever $P\ne Q$ \citep{gneiting2007strictly}. Propriety is the condition that reporting
one's true belief is never penalised relative to reporting anything else;
strict propriety is the stronger condition that reporting one's true belief is
the unique optimum. Both notions are stated for a general distribution $Q$,
not only for the one-hot distributions that arise from a single realised
class, and we use them at that level of generality throughout.

The setting of interest restricts the outcome to a single realised class
$\omega\in\{1,\dots,c\}$, represented by its one-hot indicator $y$, with
$y_\omega=1$ and $y_j=0$ otherwise. A forecast is a vector
$p\in\Delta_c=\{p\in[0,1]^c:\sum_j p_j=1\}$, the probability simplex over the
$c$ classes. Two scoring rules recur throughout: the Brier score and the
logarithmic score. We write both in \textbf{sum form},
\begin{equation}S_{BS}(p,\omega)=\sum_{j=1}^c(p_j-y_j)^2=(1-p_\omega)^2+\sum_{j\ne\omega}p_j^2,
\qquad S_{LL}(p,\omega)=-\log p_\omega,\end{equation}
and we use the sum form exclusively for the remainder of the paper. The mean
form, $\frac1c\sum_j(p_j-y_j)^2$, differs from $S_{BS}$ by a factor of $c$; we
note this here because the distinction recurs, and return to why it matters
when we relate our results to the published rules.

$S_{LL}$ is finite only where $p_\omega>0$, so every result stated for the
logarithmic base is a statement about $p$ in the relative interior of the
simplex, $\Delta_c^\circ=\{p\in\Delta_c:p_j>0\text{ for all }j\}$, not the
full closed $\Delta_c$ the Brier-base results cover. $S_{BS}$ itself is
finite and well defined on all of $\Delta_c$, including its boundary. We
state this restriction here, once, rather than repeat it at every
logarithmic-base proposition. Whether the family's logarithmic-base
guarantees extend to forecasts on the boundary ($p_j=0$ for some $j$) by a
continuity or limiting argument is not settled by anything in this paper;
we flag the gap rather than assume it away.

Following source Definition 5 \citep{ahmadian2025superior}, we partition the set of forecasts, for a fixed true class $\omega$,
into those under which $\omega$ counts as the predicted class and those under
which it does not. Write $M(p)=\{j:p_j=\max_k p_k\}$ for the argmax set of a
forecast $p$. When $M(p)$ is a singleton, $\omega$ is predicted exactly when
$\omega=\arg\max_j p_j$, and the partition is unambiguous. When $M(p)$ is not
a singleton — an exact tie among the largest forecast components — the
partition depends on a choice of convention, and we fix one explicitly rather
than leave it implicit. We distinguish three conventions, used throughout the
paper by these names.

\textbf{Definition (Convention A, permissive).} $\omega$ counts as predicted
whenever $\omega\in M(p)$: every class tied for the top spot is credited as
correct.

\textbf{Definition (Convention B, single-valued).} Correctness is decided by a
fixed single-valued rule $\delta:\Delta_c\to\{1,\dots,c\}$ with
$\delta(p)\in M(p)$, required to be induced by one fixed total order
$\prec$ on $\{1,\dots,c\}$ applied consistently across every tie-set (e.g.
$\delta(p)$ is always the $\prec$-least member of $M(p)$); $\omega$ counts
as predicted exactly when $\delta(p)=\omega$. Exactly one class is credited
as correct on every tie, chosen by the order $\prec$ rather than by $\omega$.

\textbf{Definition (Convention C, ties-to-$\xi$).} $\omega$ counts as predicted
only if $p_\omega$ strictly exceeds every other coordinate: $\omega$ is
credited as correct exactly when $M(p)=\{\omega\}$, so every exact tie
counts as incorrect regardless of which class is true.

\textbf{Example.} Take $c=3$ and the forecast $p=(0.5,0.5,0)$, so
$M(p)=\{1,2\}$, and fix the order $1\prec2\prec3$ for Convention B. Table~\ref{tab:preliminaries-1}
gives whether each true class $\omega$ counts as correctly predicted under
each convention.

\begin{table}[htbp]
\centering
\caption{Whether $p=(0.5,0.5,0)$ counts as a correct prediction of the true class $\omega$, by convention.}
\label{tab:preliminaries-1}
\begin{tabularx}{\textwidth}{Y>{\centering\arraybackslash}p{1.02in}>{\centering\arraybackslash}p{1.02in}>{\centering\arraybackslash}p{1.02in}}
\hline
Convention & $\omega=1$ & $\omega=2$ & $\omega=3$ \\
\hline
A (permissive) & correct & correct & incorrect \\
B (single-valued, $1\prec2\prec3$) & correct & incorrect & incorrect \\
C (ties-to-$\xi$) & incorrect & incorrect & incorrect \\
\hline
\end{tabularx}
\end{table}

Convention A credits both tied classes; Convention B credits exactly one,
picked by the order rather than by which class is true; Convention C
credits neither, since neither $p_1$ nor $p_2$ strictly exceeds the other.
We write $\psi$ for the set of forecasts under which $\omega$ counts as
predicted and $\xi$ for its complement, under whichever convention is in
force; where the convention matters we write $\psi_A,\xi_A$, $\psi_B,\xi_B$,
and $\psi_C,\xi_C$. The released implementation accompanying the source
paper adopts Convention A, resolving an exact tie in favour of the correct
class.


\section{Related work}
\label{sec:related-work}

We organise related work by argument rather than chronology, along three
threads: proper scoring rules and their decompositions; decision-theoretic
and cost-sensitive evaluation; and model selection and early-stopping
criteria, where our own contribution lands.

\subsection{Proper scoring rules and their decompositions}
\label{sub:related-work-proper-scoring-rules-and-their-decompositions}

Propriety and strict propriety, the framing we use throughout, follow
\citep{gneiting2007strictly}. \citep{waghmare2025proper} is a recent,
comprehensive review of the same foundations — general characterisation
results and the major families of scoring rules — and we point to it rather
than to earlier, narrower overviews \citep{carvalho2016overview} for the
general theory; its scope is continuous and distributional forecasting
rather than classification, so it grounds our framing without overlapping
our specific contribution. The Brier score itself originates with
\citep{brier1950verification}; \citep{dawid2014theory} and
\citep{landes2015probabilism} give further foundational and philosophical
treatments, the latter in this paper's own target venue.

Closer to our setting, \citep{kull2015novel} decompose proper scoring rules
for classification into calibration, refinement, adjustment, and grouping
terms, aimed at improving classifiers rather than at the evaluation-time
penalty we study. \citep{resin2023classification} extends elicitability
results to bridge single-class predictions and full predictive
distributions. Both work on the same object — a proper scoring rule for
classification — through decomposition or extension rather than through the
interval-separation characterisation and infimal-penalty question we pursue.
\citep{hoessly2026misconceptions} documents
that the Brier score is widely misinterpreted in practice — an adjacent but
different concern from ours, which is that the \textit{penalty} term added to it
has not been treated as a free parameter needing its own calibration.

A further body of work studies proper scoring rules for continuous and
ensemble forecasts specifically — calibration and sharpness
\citep{gneiting2007probabilistic}, ensemble model output statistics
\citep{gneiting2005calibrated}, CRPS decomposition
\citep{hersbach2000decomposition}, forecast verification generally
\citep{wilks2011forecast}, and robustness to scale and outliers
\citep{bolin2023local} — which we do not draw on directly, since our setting
is single-label multi-class classification rather than continuous
prediction. \citep{hu2025choice} studies scoring a target property as a weighted
sum of proper scoring rules for sub-properties — structurally the same kind
of object as our own weighted-mixture baseline (Section~\ref{sub:results-baselines-does-the-frontier-require-superiority-or-only-re-weighting}),
though aimed at preserving propriety through the weighting rather than, as
our baseline deliberately does, isolating what a non-proper weighted
combination achieves on its own.

\subsection{Decision-theoretic and cost-sensitive evaluation}
\label{sub:related-work-decision-theoretic-and-cost-sensitive-evaluation}

\citep{good1952rational} connects probability estimation to rational
decision-making and is the earliest work we draw on for the decision-theoretic
grounding of a scoring rule. \citep{seidenfeld2012forecasting} extends proper
scoring to imprecise, interval-valued probability forecasts. This thread is
thinner in the existing literature than the other two, and recent work
closes much of the gap. \citep{shahroudi2025aligning} learn a proxy
evaluation function, via a neural-network-weighted proper scoring rule, that
aligns an upstream metric with downstream decision value, for continuous
prediction tasks. This addresses the same underlying mismatch we do — a
metric's free parameter set without reference to downstream use — through
adaptive weighting rather than a closed-form infimal penalty, and for
regression rather than classification. \citep{schneider2026decision}
formalise \textit{decision-alignment} directly and show that coordinate-dependent
metrics such as expected calibration error cannot be decision-aligned with
pointwise-separable utilities, a result that bears on how any calibration
diagnostic, including the ones in our own experimental design, should be
read. \citep{lombardo2025cost} propose a reweighting framework for
cost-sensitive binary classification evaluation, an alternative to
resampling rather than a propriety-preserving penalty.

\subsection{Model selection and early-stopping criteria}
\label{sub:related-work-model-selection-and-early-stopping-criteria}

\citep{prechelt2002early} shows that the early-stopping criterion is usually
chosen without a systematic basis, and quantifies the tradeoff between
stopping patience and generalisation. Our own $\lambda$-selection problem is
the same tradeoff restated for the scoring rule doing the stopping, rather
than for the patience parameter itself. \citep{li2020gradient} gives a
theoretical account of why early stopping is robust to label noise in
overparameterised networks, directly relevant to the noisy and hard regimes
in our experimental design. \citep{guo2017calibration} and
\citep{vaicenavicius2019evaluating} establish that modern networks are
poorly calibrated and develop a general framework for evaluating that
calibration. \citep{shao1993linear} shows, in the classical linear-model
setting, that a common model-selection procedure is asymptotically
inconsistent — a caution about model-selection criteria generally rather
than a result about scoring rules specifically.

Most recently, checkpoint- and stopping-criterion design has moved toward
signals that avoid a held-out validation set or a fixed scalar score
altogether: \citep{jamshidi2025} stop training using only gradient
information via posterior sampling, and \citep{nguyen2025uncertainty} select
checkpoints by per-sample uncertainty on hard examples. Both are orthogonal
to our question: they address whether a
validation-based score is needed at all, whereas we ask, given that a
validation-based scoring rule is being used for exactly this purpose, how
its penalty term should be set.

Table~\ref{tab:related-work-1} collects the closest of these along the three dimensions our own
contribution combines: whether the work provides a pointwise ordering
guarantee on individual predictions, whether it preserves propriety, and
whether it yields a closed-form penalty threshold.

\begin{table}[htbp]
\centering
\caption{Positioning against the closest prior approaches, on the dimensions this paper's own contribution combines.}
\label{tab:related-work-1}
\begin{tabularx}{\textwidth}{YYYYY}
\hline
Work & What it targets & Pointwise ordering guarantee & Propriety preserved & Closed-form penalty threshold \\
\hline
\citet{kull2015novel} & Decomposes a proper rule into calibration, refinement, adjustment, and grouping terms & Not addressed & N/A (decomposes an already-proper rule) & No \\
\citet{resin2023classification} & Elicitability bridge, single-class to full predictive distributions & Not addressed & Not the focus & No \\
\citet{hu2025choice} & Weighted sum of proper scoring rules for sub-properties & Not addressed & Weighting is designed to preserve it & No \\
\citet{shahroudi2025aligning} & Learned proxy aligning an upstream metric with downstream decision value (regression) & Not addressed & Not the focus & No (adaptive weighting) \\
\citet{schneider2026decision} & Formal decision-alignment analysis; shows ECE-like metrics cannot align with pointwise-separable utilities & Diagnostic result, not a constructive guarantee & Not the focus & No \\
\citet{lombardo2025cost} & Reweighting framework for cost-sensitive binary classification & Not addressed & Not propriety-preserving & No \\
This paper & Penalty magnitude for a decision-aware scoring rule, multi-class classification & Yes (superiority, theory part 2) & Preserved (Theorem~\ref{thm:strict-propriety}) & Yes ($\lambda^\ast$, Corollary~\ref{cor:closed-forms}) \\
\hline
\end{tabularx}
\end{table}

This is where our own contribution lands: none of the above treats the
penalty magnitude of a decision-aware scoring rule as a free parameter with
its own closed-form infimum, set from a characterisation of the rule's
range over correct and incorrect predictions rather than from one endpoint
of that range alone.


\section{Superiority as interval separation}
\label{sec:theory-1}

\subsection{Superiority, restated}
\label{sub:theory-1-superiority-restated}

Source Definition 5 \citep{ahmadian2025superior} calls a scoring rule $S$ \textit{superior} if every correctly classified forecast
scores strictly better than every incorrectly classified one, for every pair
of realised classes: for all $p^{(1)}\in\psi$ with true class $\omega^{(1)}$
and all $p^{(2)}\in\xi$ with true class $\omega^{(2)}$,
\begin{equation}S(p^{(1)},\omega^{(1)})<S(p^{(2)},\omega^{(2)}).\end{equation}
Both $S_{BS}$ and $S_{LL}$ are invariant under simultaneous relabelling of the
classes in $p$ and $\omega$, so $\sup_\psi S$ and $\inf_\xi S$ do not depend on
$\omega$ as values, and we may fix $\omega=i$ throughout when computing them.
This symmetry does not extend to \textit{attainment} under Convention B: a fixed
order $\prec$ is not relabelling-invariant, so which extremum is attained can
depend on where $i$ sits in $\prec$. We return to this point below. Every
proof that fixes $\omega=i$ "by symmetry" asserts only the value of an
extremum, never its attainment.

\subsection{The characterisation}
\label{sub:theory-1-the-characterisation}

Superiority, quantified over all pairs $(p^{(1)},p^{(2)})$, is equivalent to a
statement about two numbers: the supremum of $S$ over $\psi$ and the infimum
of $S$ over $\xi$. Write $\Sigma_\psi=\sup_{p\in\psi}S(p,\omega)$ and
$\Sigma_\xi=\inf_{p\in\xi}S(p,\omega)$.

\begin{proposition}
\label{prop:characterisation}
\textit{(i) If $\Sigma_\psi<\Sigma_\xi$, then $S$ is superior. (ii)
If $\Sigma_\psi>\Sigma_\xi$, then $S$ is not superior. (iii) If
$\Sigma_\psi=\Sigma_\xi$, then $S$ is superior if and only if at least one of
the two extrema is not attained.}
\end{proposition}

\textit{Proof.} (i) For any $p^{(1)}\in\psi$, $p^{(2)}\in\xi$,
$S(p^{(1)},\omega^{(1)})\le\Sigma_\psi<\Sigma_\xi\le S(p^{(2)},\omega^{(2)})$.

(ii) Since $\Sigma_\psi>\Sigma_\xi$ and $\Sigma_\psi$ is a supremum, choose
$p^{(1)}\in\psi$ with $S(p^{(1)},\omega^{(1)})>\Sigma_\xi$; since $\Sigma_\xi$
is an infimum and $\Sigma_\xi<S(p^{(1)},\omega^{(1)})$, choose $p^{(2)}\in\xi$
with $S(p^{(2)},\omega^{(2)})<S(p^{(1)},\omega^{(1)})$. This pair violates the
superiority inequality.

(iii) If both extrema are attained, at $p^{(1)}_\star$ and $p^{(2)}_\star$,
then $S(p^{(1)}_\star,\cdot)=\Sigma_\psi=\Sigma_\xi=S(p^{(2)}_\star,\cdot)$ and
the strict inequality required for superiority fails for that pair. If at
least one extremum is not attained, then for every admissible pair,
$S(p^{(1)},\cdot)\le\Sigma_\psi=\Sigma_\xi\le S(p^{(2)},\cdot)$ has at least
one strict inequality, so the pair satisfies superiority. $\square$

Proposition~\ref{prop:characterisation} is a restatement of the definition of superiority rather than a
deep result: it replaces a quantifier over pairs of forecasts with two
optimisation problems in one variable. Its value is that both $\Sigma_\psi$
and $\Sigma_\xi$ admit closed forms, which we compute next. It is also the
reason a tie convention must be fixed at all, since case (iii) is exactly
where the convention decides the outcome.

\subsection{The four endpoints}
\label{sub:theory-1-the-four-endpoints}

\begin{proposition}[supremum over $\psi$, Brier]
\label{prop:sup-psi-brier}
\textit{For $S_{BS}$,
$\Sigma_\psi=\dfrac{c-1}{c}$ under both conventions. Attained, uniquely at the
uniform forecast, under Convention A for every $i$; under order-induced
Convention B, attained if and only if $i=j_0$.}
\end{proposition}

\textit{Proof.} Fix $\omega=i$ and let $p\in\psi_A$, $t=p_i$, so $p_j\le t$ for all
$j\ne i$. Summing over all $j$, $1=\sum_j p_j\le ct$, so $t\ge\frac1c$. Since
$0\le p_j\le t$ for $j\ne i$, $p_j^2\le tp_j$, so
$\sum_{j\ne i}p_j^2\le t\sum_{j\ne i}p_j=t(1-t)$. Hence
\begin{equation}S_{BS}(p,i)=(1-t)^2+\sum_{j\ne i}p_j^2\le(1-t)^2+t(1-t)=1-t\le1-\tfrac1c=\tfrac{c-1}{c}.\end{equation}
Equality throughout requires $t=\frac1c$ and $p_j\in\{0,t\}$ for every $j\ne
i$ with the terms summing to $1-\frac1c$, which forces every $p_j=\frac1c$:
the unique maximiser is the uniform forecast $u:=(\frac1c,\dots,\frac1c)$,
which lies in $\psi_A(i)$ since every class is in $M(p)$. This proves the
Convention A claim.

For Convention B, fix $i$ and let $p\in\psi_B(i)$, so $\delta(p)=i$; write
$k=|M(p)|\in\{1,\dots,c\}$ and $t=p_i=\max_jp_j$. The $k-1$ other classes in
$M(p)$ also equal $t$, and the remaining $c-k$ classes are excluded from
$M(p)$ by definition, so $p_j<t$ \textit{strictly} for each; write $R:=1-kt$ for
their sum. Since $p_j<t$, also $p_j^2<tp_j$ whenever $p_j>0$ and $p_j^2\le
tp_j$ when $p_j=0$, so their squared mass is at most $tR$ as above. Hence
\begin{equation}S_{BS}(p,i)=(1-t)^2+(k-1)t^2+\!\!\sum_{j\notin M(p)}\!\!p_j^2\ \le\
(1-t)^2+(k-1)t^2+t(1-kt)=1-t,\end{equation}
the same bound as under Convention A, holding at every tie-set size $k$, not
only $k=1$ or $k=c$. Summing all $c$ coordinates, $kt+R=1$. If $k<c$ there is
at least one free coordinate, each strictly below $t$, so $R<(c-k)t$
strictly, giving $kt+R<ct$, i.e. $1<ct$, i.e. $t>\frac1c$ strictly. So
$t=\frac1c$ — needed for the bound to reach $\frac{c-1}{c}$ — is only
possible when $k=c$, the full tie with no free coordinates at all, forcing
$p=u:=(\frac1c,\dots,\frac1c)$. Since $M(u)=\{1,\dots,c\}$, $\delta(u)=j_0$ by
definition of an order-induced tie-break on a full tie.

\textbf{If $i=j_0$:} $M(u)=\{1,\dots,c\}$, so $\delta(u)=j_0$ by definition —
$u\in\psi_B(j_0)$, and $S_{BS}(u,j_0)=\frac{c-1}{c}$ is \textbf{attained}.

\textbf{If $i\ne j_0$:} $\delta(u)=j_0\ne i$, so $u\notin\psi_B(i)$ — and $u$ was
the only point, at the only tie-set size ($k=c$) that could reach the
bound. The supremum is \textbf{not attained}. It is still the supremum: take
$k=1$ (no tie at all), $p_i=\tfrac1c+\varepsilon$,
$p_j=\tfrac1c-\tfrac{\varepsilon}{c-1}$ for $j\ne i$, so $M(p)=\{i\}$ and
$\delta(p)=i$ for every $\prec$ (nothing to break); $S_{BS}(p,i)\to
\frac{c-1}{c}$ as $\varepsilon\to0^+$. $\square$


This matches source Theorem 3, whose argument sets $\varepsilon=0$ to
simplify — exactly the tie case, which is where the attainment question we
make explicit here originates.

\begin{proposition}[infimum over $\xi$, Brier]
\label{prop:inf-xi-brier}
\textit{For $c>2$, $S_{BS}$,
$\Sigma_\xi=\dfrac12$, independent of $c$ and of $i$. The minimiser is a
two-way tie, $p_i=p_k=\frac12$ for one $k\ne i$ with all other coordinates
zero.}
\end{proposition}

\textit{Proof.} Since $\xi_A\subseteq\xi_B\subseteq\overline{\xi_A}$ for any
single-valued $\delta$ with $\delta(p)\in M(p)$, and $S_{BS}$ is continuous on
the compact set $\Delta_c$, the value of the infimum coincides on $\xi_A$,
$\xi_B$, and the closure $\overline{\xi_A}=\{p:\exists k\ne i,\,p_k\ge p_i\}$;
we compute it over $\overline{\xi_A}$, and treat attainment separately. By
symmetry over the coordinates $j\ne i$, fix a binding class $k$ and write
$t=p_i$, with constraint $p_k\ge t$.

For fixed $t$, minimise $\sum_{j\ne i}p_j^2$ subject to $\sum_{j\ne i}p_j=1-t$,
$p_j\ge0$, $p_k\ge t$. A sum of squares with fixed total is minimised by
equalising the terms; the unconstrained optimum $p_j=(1-t)/(c-1)$ for every
$j\ne i$ satisfies $p_k\ge t$ iff $t\le1/c$, giving
\begin{equation}m(t)=\frac{(1-t)^2}{c-1}\quad(0\le t\le\tfrac1c).\end{equation}
For $t>1/c$ the constraint binds; strict convexity of the objective on a
convex feasible set makes the minimiser unique, with $p_k=t$ at its lower
bound and the remaining mass $1-2t$ spread equally over the $c-2$ other
classes:
\begin{equation}m(t)=t^2+\frac{(1-2t)^2}{c-2}\quad(\tfrac1c<t\le\tfrac12),\end{equation}
and $\overline{\xi_A}$ is empty for $t>\frac12$, since $p_i+p_k\ge2t>1$ would
contradict $\sum_jp_j=1$. The two branches agree at $t=1/c$.

Set $f(t)=(1-t)^2+m(t)$. On $[0,\frac1c]$,
$f_1(t)=\frac{c}{c-1}(1-t)^2$ is strictly decreasing, so
$\min_{[0,1/c]}f_1=f_1(1/c)=\frac{c-1}{c}$. On $(\frac1c,\frac12]$,
\begin{equation}f_2(t)=(1-t)^2+t^2+\frac{(1-2t)^2}{c-2},\end{equation}
\begin{equation}f_2''(t)=4+\frac{8}{c-2}>0,\end{equation}
so $f_2$ is strictly convex, with stationary point solving
$-2(1-t)+2t-\frac{4(1-2t)}{c-2}=0$, i.e. $4tc-2c=0$, giving $t=\frac12$ for
every $c$. Since the stationary point is the right endpoint, $f_2$ is
decreasing on $(\frac1c,\frac12]$ and
$\min_{(1/c,1/2]}f_2=f_2(\frac12)=\frac14+\frac14+0=\frac12$. Comparing the two
branch minima, $\frac{c-1}{c}>\frac12\iff c>2$, so for $c>2$ the second branch
wins and $\min f=\frac12$, attained at $t=\frac12$. Strict convexity forces
every minimising configuration to be the two-way tie $p_i=p_k=\frac12$, other
coordinates zero. $\square$

Attainment of $\Sigma_\xi$ is where the tie convention enters, and it is here
that fixing $\omega=i$ "by symmetry" stops being innocent. Under Convention A
the minimising tie lies in $\psi_A$ (permissively correct), so the infimum
over $\xi_A$ is approached but never attained, for every $i$. Under
order-induced Convention B, write $j_0$ for the $\prec$-least class. If
$i\ne j_0$, the tie $p_i=p_{j_0}=\frac12$ resolves to $\delta(p)=j_0\ne i$,
placing this point in $\xi_B$: the infimum is attained. If $i=j_0$, the class
$j_0$ wins every tie it participates in, so no two-way tie resolves away from
it, and the infimum is only approached, exactly as under Convention A. In
both sub-cases, exactly one of $\Sigma_\psi,\Sigma_\xi$ is attained — never
both, never neither — and which one flips with $i$'s position in $\prec$; we
state this dichotomy in general form as Lemma~\ref{lem:attainment-dichotomy} below, since it is what the
minimality theorem in the next section relies on directly, not the value
$\frac12$ by itself.

\begin{proposition}[supremum over $\psi$, logarithmic; $p\in\Delta_c^\circ$]
\label{prop:sup-psi-log}
\textit{$\Sigma_\psi=\log c$. Attained under Convention A; under Convention B,
attained if and only if $i=j_0$.}
\end{proposition}

\textit{Proof.} $S_{LL}(p,i)=-\log p_i$ is strictly decreasing in $p_i$, so the
supremum is attained by minimising $p_i$. Under Convention A,
$p\in\psi_A$ gives $p_j\le p_i$ for all $j$, hence $1=\sum_jp_j\le cp_i$ and
$p_i\ge1/c$, so $S_{LL}(p,i)\le\log c$, with equality iff $p_i=1/c$, which (as
in Proposition~\ref{prop:sup-psi-brier}) forces $p$ uniform; the uniform vector lies in $\psi_A$, so
the supremum is attained.

Under order-induced Convention B, the unique minimiser of $p_i$ subject to
$p_i=\max_jp_j$ is again the uniform vector, and $\delta(\text{uniform})=j_0$.
If $i=j_0$, the uniform vector lies in $\psi_B(j_0)$ and the supremum is
attained exactly as under Convention A. If $i\ne j_0$, $\psi_B(i)$ excludes
the uniform vector and, by uniqueness of the minimiser, no other point
achieves $p_i=1/c$ while remaining in $\psi_B(i)$; so $p_i>1/c$ strictly on
$\psi_B(i)$, $S_{LL}(p,i)<\log c$ throughout, and the supremum is approached
along $p_i=1/c+\varepsilon$ but not attained. $\square$

This matches source Theorem 4.

\begin{proposition}[infimum over $\xi$, logarithmic; $p\in\Delta_c^\circ$]
\label{prop:inf-xi-log}
\textit{$\Sigma_\xi=\log 2$, independent of $c$. Attained under Convention B if and
only if $i\ne j_0$; never attained under Convention A.}
\end{proposition}

\textit{Proof.} $S_{LL}(p,i)=-\log p_i$ is minimised by maximising $p_i$. Since
$\xi_A\subseteq\xi_B\subseteq\overline{\xi_A}$ for any single-valued $\delta$,
the value bound holds regardless of $\delta$: for $p\in\xi_B$ there is $k\ne
i$ with $p_k\ge p_i$, so $1=\sum_jp_j\ge p_i+p_k\ge2p_i$, giving $p_i\le
\frac12$ and $S_{LL}(p,i)\ge\log2$, with equality iff $p_i=p_k=\frac12$ and
all other coordinates zero. Under order-induced Convention B with $j_0$ as
above: if $i\ne j_0$, the tie $p_i=p_{j_0}=\frac12$ resolves to
$\delta(p)=j_0\ne i$, so this point lies in $\xi_B(i)$ and the infimum is
attained; if $i=j_0$, no two-way tie resolves away from $j_0$, so
$\xi_B(j_0)=\xi_A(j_0)$ and the infimum is only approached. Under Convention
A the requirement $p_k>p_i$ gives $p_i<\frac12$ strictly, so $S_{LL}(p,i)>
\log2$ throughout and the infimum is approached but never attained. $\square$

This matches source Theorem 4's counterpart for the lower endpoint, which the
source paper does not state; the bound uses only two coordinates and so, as
with Proposition~\ref{prop:inf-xi-brier}, does not depend on $c$.

\begin{lemma}[attainment dichotomy]
\label{lem:attainment-dichotomy}
\textit{Fix $\omega=i$ and an order-induced
$\delta$ with $j_0=\arg\min_\prec\{1,\dots,c\}$. For both $S_{BS}$ and
$S_{LL}$, exactly one of $\Sigma_\psi,\Sigma_\xi$ is attained: under
Convention A, $\Sigma_\psi$ is attained and $\Sigma_\xi$ is not, for every
$i$; under Convention B, $\Sigma_\psi$ is attained and $\Sigma_\xi$ is not
when $i=j_0$, and the reverse when $i\ne j_0$.}
\end{lemma}

\textit{Proof.} This is the content of Propositions~\ref{prop:sup-psi-brier}--\ref{prop:inf-xi-log} above, collected: the two
extremisers in each case are ties (the uniform vector, and the pairwise
$\frac12$-tie), so their fate under an order-induced $\delta$ turns only on
where $i$ sits relative to $j_0$, and the same argument applies to both
scoring rules since it uses nothing beyond "the extremiser is a tie." In
every row exactly one column holds. $\square$

\begin{figure}[htbp]
\centering
\includegraphics[width=3.0in]{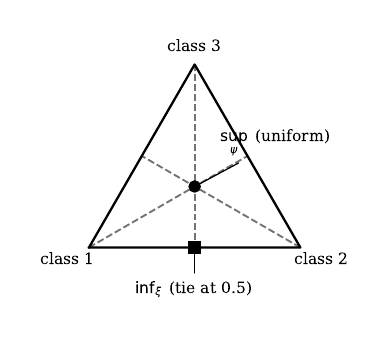}
\caption{The probability simplex at $c=3$. Dashed lines mark the three
argmax tie loci $\{p:p_i=p_j\}$. The uniform vector attains
$\sup_\psi S_{BS}=0.667$ (Proposition~\ref{prop:sup-psi-brier}); a pairwise tie at the value $0.5$
attains $\inf_\xi S_{BS}=0.500$ (Proposition~\ref{prop:inf-xi-brier}). Both marked points sit
exactly
on the tie set, a measure-zero subset of the simplex — precisely where the
tie conventions disagree with each other about attainment.}
\label{fig:simplex}
\end{figure}

Figure~\ref{fig:simplex} makes Lemma~\ref{lem:attainment-dichotomy} concrete at $c=3$: both extremisers
sit on the tie set, which is exactly where an order-induced $\delta$ can
disagree with the permissive convention about which one is attained.

\subsection{The minimiser and the source paper's own example}
\label{sub:theory-1-the-minimiser-and-the-source-paper-s-own-example}

The minimiser of Proposition~\ref{prop:inf-xi-brier} is a two-way tie with the true class and one
incorrect class each carrying probability mass $\frac12$, independent of $c$.
This is the limiting form of the source paper's own motivating example, the
vector $B=[0.51,0.49,0]$, whose sum-form Brier score is $0.520$ against the
bound $\frac12$ established here. The source paper's own worked example
points directly at the quantity Proposition~\ref{prop:inf-xi-brier} computes; it does not compute
it.

\subsection{The $c=2$ corollary}
\label{sub:theory-1-the-c-2-corollary}

\begin{corollary}
\label{cor:c2}
\textit{At $c=2$, $\lambda^\ast:=\Sigma_\psi-\Sigma_\xi=0$ for both
$S_{BS}$ and $S_{LL}$.}
\end{corollary}

\textit{Proof.} For $S_{BS}$, Proposition~\ref{prop:inf-xi-brier}'s second branch is vacuous at $c=2$;
direct minimisation of $(1-t)^2+t^2$ over $t\le\frac12$ gives $\frac12$ at
$t=\frac12$, coinciding with $\Sigma_\psi=\frac{c-1}{c}=\frac12$. For
$S_{LL}$, $\Sigma_\psi=\log2=\Sigma_\xi$ directly from Propositions~\ref{prop:sup-psi-log} and~\ref{prop:inf-xi-log}.
$\square$

Corollary~\ref{cor:c2} recovers the source paper's restriction to $c>2$ as a special
case of the general closed forms established above, rather than as a
separately argued condition.


\section{The penalised family and its propriety}
\label{sec:theory-2}

\subsection{The family}
\label{sub:theory-2-the-family}

For a strictly proper base rule $S_{\text{base}}$ and $\lambda\ge0$, define
\begin{equation}S_\lambda(p,\omega)=S_{\text{base}}(p,\omega)+\lambda\cdot\mathbb{1}[p\in\xi].\end{equation}
The published $PBS$ and $PLL$ (source Eq.\ 42 and Eq.\ 43) are the members of
this family at, respectively, the Brier base with $\lambda=\frac{c-1}{c}$ and
the logarithmic base with $\lambda=\log c$; we return to this positioning
below. The construction generalises to a full misclassification cost matrix
in place of the single scalar $\lambda$, with propriety plausibly preserved
by the same argument. We do not establish this here and treat it only as a
direction in the discussion.

\subsection{Propriety of the penalty term}
\label{sub:theory-2-propriety-of-the-penalty-term}

Under Convention B, the penalty is the $0$–$1$ loss of the single-valued
decision $\delta(p)\in M(p)$:
\begin{equation}L_B(p,\omega)=\mathbb{1}[\delta(p)\ne\omega].\end{equation}
Under Convention A it is instead $L_A(p,\omega)=\mathbb{1}[\omega\notin
M(p)]$, which is not of the form $\mathbb{1}[\delta(p)\ne\omega]$ once
$|M(p)|>1$. This distinction is the entire content of Proposition~\ref{prop:convention-a-failure} below.

\begin{lemma}
\label{lem:penalty-proper}
\textit{$L_B$ is proper but not strictly proper.}
\end{lemma}

\textit{Proof.} For any $Q\in\Delta_c$,
\begin{equation}\mathbb{E}_Q[L_B(p,\cdot)]=\sum_jQ_j\mathbb{1}[\delta(p)\ne j]=1-Q_{\delta(p)}\ge1-\max_jQ_j.\end{equation}
Taking $p=Q$ gives $\delta(Q)\in M(Q)=\arg\max Q$, hence
$\mathbb{E}_Q[L_B(Q,\cdot)]=1-\max_jQ_j$, the lower bound just derived, so
$L_B$ is proper. It is not strictly proper: every $p$ with
$\delta(p)\in\arg\max Q$ attains the same value. $\square$

The argument does not require $\arg\max Q$ to be a singleton — $\max_jQ_j$ is
well defined regardless, and $\delta(Q)$ selects some element of it — so
tied modes in $Q$ do not obstruct propriety of the penalty term. Lemma~\ref{lem:penalty-proper} also
needs no order-inducedness hypothesis on $\delta$: the proof evaluates
$\delta$ only at the single point $p=Q$, and never compares how $\delta$
treats two different tie-sets against each other. The order-inducedness
hypothesis that Lemma~\ref{lem:attainment-dichotomy} (Section~\ref{sec:theory-1}) requires is specific to the
superiority boundary. Propriety of the penalty term is robust to which
single-valued $\delta$ is chosen.

\begin{lemma}
\label{lem:sum-strictly-proper}
\textit{If $S_1$ is strictly proper, $S_2$ is proper, and $\lambda\ge0$,
then $S_1+\lambda S_2$ is strictly proper.}
\end{lemma}

\textit{Proof.} For $P\ne Q$, strict propriety of $S_1$ gives
$\mathbb{E}_Q[S_1(P,\cdot)]>\mathbb{E}_Q[S_1(Q,\cdot)]$, and propriety of
$S_2$ gives $\lambda\,\mathbb{E}_Q[S_2(P,\cdot)]\ge\lambda\,\mathbb{E}_Q[S_2(Q,\cdot)]$.
Adding, $\mathbb{E}_Q[(S_1+\lambda S_2)(P,\cdot)]>\mathbb{E}_Q[(S_1+\lambda
S_2)(Q,\cdot)]$. $\square$

\begin{theorem}
\label{thm:strict-propriety}
\textit{Under Convention B, $S_\lambda=S_{\text{base}}+\lambda L_B$ is
strictly proper for every $\lambda\ge0$, for any strictly proper base.}
\end{theorem}

\textit{Proof.} Lemma~\ref{lem:penalty-proper} gives propriety of $L_B$; Lemma~\ref{lem:sum-strictly-proper} applied with
$S_1=S_{\text{base}}$, $S_2=L_B$ gives the result. $\square$

Theorem~\ref{thm:strict-propriety} is the corrected argument for source Theorem 5: the theorem itself
is true, and it holds for the whole family rather than for the two published
instances alone. Lemma~\ref{lem:sum-strictly-proper} by itself is an elementary fact about convex
combinations of proper and strictly proper rules; the substantive content
here is Lemma~\ref{lem:penalty-proper} — that the single-valued mode-mismatch loss $L_B$ stays
proper regardless of how $Q$'s own mode is tied — and, below, correctly
diagnosing which convention the argument actually needs. We return to the
published argument in the next subsection.

\textbf{Remark 1 (a second route).} The expected score of a truthful forecaster is
\begin{equation}G_\lambda(Q)=\mathbb{E}_Q[S_\lambda(Q,\cdot)]=G_{\text{base}}(Q)+\lambda\big(1-\max_jQ_j\big).\end{equation}
$G_{\text{base}}$ is concave, the standard characterisation of propriety
\citep{gneiting2007strictly}, and $Q\mapsto-\max_jQ_j$ is
concave because $\max$ is convex; a non-negative combination of concave
functions is concave. This offers an independent confirmation of Theorem~\ref{thm:strict-propriety}
through the generalised-entropy representation, rather than a second proof —
the representation theorem underlying it requires regularity conditions this
remark does not check.

\subsection{Convention A does not preserve strict propriety}
\label{sub:theory-2-convention-a-does-not-preserve-strict-propriety}

\begin{proposition}
\label{prop:convention-a-failure}
\textit{Under Convention A, a forecaster who reports the uniform
vector incurs expected penalty zero under every $Q$, so $S_\lambda$ is not
strictly proper.}
\end{proposition}

\textit{Proof.} Under Convention A, $\mathbb{E}_Q[L_A(p,\cdot)]=1-\sum_{j\in
M(p)}Q_j$. The uniform vector $u=(\frac1c,\dots,\frac1c)$ has $M(u)=\{1,\dots,c\}$,
so $\sum_{j\in M(u)}Q_j=1$ and the expected penalty is zero for every $Q$. A
truthful forecaster with a unique mode instead pays $\lambda(1-\max_jQ_j)>0$
whenever $\lambda>0$. Hence the uniform report beats the truthful report
whenever $\lambda$ exceeds
\begin{equation}\lambda_A(Q):=\frac{\mathbb{E}_Q[S_{\text{base}}(u,\cdot)]-G_{\text{base}}(Q)}{1-\max_jQ_j}.\end{equation}
$\square$

For $c=3$, $Q=(0.5,0.3,0.2)$ with the Brier base, $\lambda_A(Q)\approx0.093$;
the published $\lambda=\frac23$ exceeds it, and the implemented effective
$\lambda=2$ exceeds it more than twentyfold. Numerically, at $c=13$ the
published $\lambda$ is exceeded on 2999 of 3000 random draws of $Q$, with
worst-case deficit $0.804$, and the implemented penalty is exceeded on every
draw. Under Convention B, by contrast, none of the same draws produce a
violation, at any of the penalty values tested, consistent with Theorem~\ref{thm:strict-propriety}.

\begin{figure}[htbp]
\centering
\includegraphics[width=3.3in]{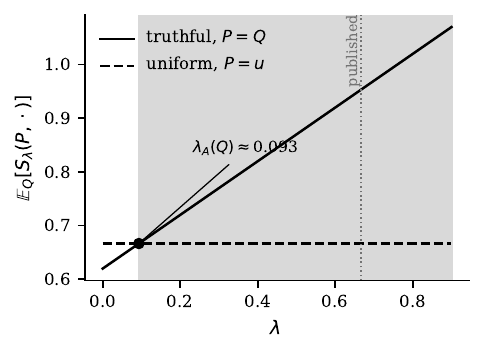}
\caption{Expected sum-form Brier-penalised score at $c=3$,
$Q=(0.5,0.3,0.2)$, under Convention A, for a truthful forecaster ($P=Q$)
and a forecaster who always reports uniform ($P=u$). Both curves are exact
evaluations of the expected score on a $\lambda$ grid, not a simulation.
The uniform report incurs zero expected penalty at every
$\lambda$ under this convention, so its curve is flat; the truthful curve
rises and crosses it at $\lambda_A(Q)\approx0.093$, found by linear
interpolation between grid points, not by evaluating a separate
closed-form formula. Shaded region: $\lambda>\lambda_A(Q)$, where
misreporting scores strictly better than truthful reporting. The published
penalty $\lambda=0.667$ (dotted) lies inside this region.}
\label{fig:propriety-failure}
\end{figure}

Figure~\ref{fig:propriety-failure} shows this worked case directly: the two
curves cross at $\lambda_A(Q)$, and the published penalty already lies past
the crossing, in the region where the uniform report wins.

Three things belong together here, in this order. The escape requires
$|M(p)|>1$, an exact argmax tie and a Lebesgue-null set of forecasts, so a
trained network with real-valued outputs almost surely never produces one and
no reported experiment is affected by it. It is nonetheless a genuine defect,
because strict propriety quantifies over all forecasts, not almost all, and
exact ties do occur in practice — at initialisation, under saturation, and in
low-precision arithmetic. The remedy is a one-line specification: adopt a
single-valued decision rule. Theorem~\ref{thm:minimality} below shows that the infimal penalty
$\lambda^\ast$ is unaffected by doing so, though the single-valued convention
changes whether $\lambda^\ast$ itself is admissible. Nothing else in the
framework moves. It is strict propriety specifically that requires the
single-valued convention, and the recommendation is only to specify a tie
rule — something the source paper had no occasion to state.

\subsection{Convention C does not preserve strict propriety either}
\label{sub:theory-2-convention-c-does-not-preserve-strict-propriety-either}

The choice of convention is genuinely a three-way trade-off, not a two-way
one. Alongside the permissive Convention A and the single-valued Convention B
already defined, a third convention, Convention C, resolves a tie by
excluding it from correctness altogether: $\omega$ counts as predicted only
if $p_\omega$ strictly exceeds every other coordinate, with every exact tie
assigned to $\xi$. Its propriety failure is narrower than Convention A's in
\textit{scope} — triggered by $Q$ itself having a
tied mode, rather than by any $Q$ once the forecast lands on a tie — but not
narrower in \textit{severity} where it occurs: at a tied-mode $Q$, Convention C
fails propriety outright, for every $\lambda>0$, with no threshold to clear;
Convention A's failure (Proposition~\ref{prop:convention-a-failure} above) needs $\lambda$ above a
threshold $\lambda_A(Q)$ first. Convention C also admits $\lambda=\lambda^\ast$
for superiority, proved below as Lemma~\ref{lem:convention-c-attainment} — by the reverse of Convention A's
attainment pattern, not the same one.

\begin{proposition}
\label{prop:convention-c-failure}
\textit{Under Convention C, the penalty
$L_C(p,\omega)=\mathbb{1}[M(p)\ne\{\omega\}]$ — zero exactly when $\omega$ is
the unique maximiser of $p$, so that every tie counts against every class,
whether or not $\omega$ is part of it — is proper but not strictly proper
whenever $Q$ has a unique mode, and fails to be proper at all whenever $Q$
has a tied mode, for every $c\ge2$.}
\end{proposition}

\textit{Proof.} Write $\mathbb{E}_Q[L_C(p,\cdot)]=\sum_jQ_j\mathbb{1}[M(p)\ne\{j\}]
=1-\sum_{j:M(p)=\{j\}}Q_j$, which equals $1-Q_k$ if $M(p)=\{k\}$ for some
$k$, and $1$ if $p$ itself has a tie — this holds for any $Q$, tied-mode or
not. If $Q$ has a unique mode, minimising over $k$ gives $1-\max_jQ_j$,
attained by every $p$ with unique maximiser at $\arg\max Q$, not only by $Q$
itself; $L_C$ is therefore proper but not strictly proper here, by the same
argument as Lemma~\ref{lem:penalty-proper}.

If instead $Q$ has a tied mode, $M(Q)$ has at least two elements, so no $k$
satisfies $M(Q)=\{k\}$ and $\mathbb{E}_Q[L_C(Q,\cdot)]=1$ exactly — the worst
possible value, regardless of the tie-set's size. Let $q^\ast=\max_jQ_j$ and
$k\in\arg\max Q$, and for $\varepsilon>0$ let $p^{(\varepsilon)}$ be $Q$ with
its $k$-th coordinate increased by $\varepsilon$ and renormalised. For small
enough $\varepsilon$, $M(p^{(\varepsilon)})=\{k\}$, so
$\mathbb{E}_Q[L_C(p^{(\varepsilon)},\cdot)]=1-p^{(\varepsilon)}_k\to1-q^\ast$ as
$\varepsilon\to0^+$, strictly below $1$ since $q^\ast>0$ always. The report
$p^{(\varepsilon)}$ therefore beats the truthful report $Q$ by a margin
approaching $q^\ast>0$, so $L_C$ is not proper. $\square$

The same construction shows the failure is not confined to the penalty term:
for any strictly proper base and any $\lambda>0$, $p^{(\varepsilon)}$ beats
$Q$ in expected $S_\lambda$ score as $\varepsilon\to0^+$, since the base
rule's strict-propriety gap between $p^{(\varepsilon)}$ and $Q$ shrinks to
zero by continuity while the penalty term's gap stays at $\lambda q^\ast>0$
throughout. So $S_\lambda$ is not proper at any tied-mode $Q$ under
Convention C, for every $\lambda>0$ — with no threshold to clear, unlike
Convention A's failure in Proposition~\ref{prop:convention-a-failure}, which requires $\lambda>\lambda_A(Q)$.

At $c=2$ the only tied-mode $Q$ is $(\frac12,\frac12)$, and $\lambda^\ast=0$
there (Corollary~\ref{cor:closed-forms} below). So — since Proposition~\ref{prop:convention-c-failure} holds for every
$\lambda>0$ and Corollary~\ref{cor:admissible-set} gives the whole admissible set under Convention B
as $(\lambda^\ast,\infty)$ — this is not a narrow failure near a boundary but
the entire admissible range: Convention C offers no usable $\lambda$ at
$c=2$ whenever $Q$ can have a tied mode.

Confirmed by 1300 closed-form checks (seed 20250815, 0 failing): generic and
tied-mode $Q$ at $c\in\{2,3,5,8,13\}$ across every tie-set size, violation of
propriety at $\lambda\in\{10^{-6},10^{-3},0.1,1,10\}$ for two different base
rules, a dedicated $c=2$ sweep from $\lambda=10^{-9}$ to $100$, and a direct
structural check that any report with a tie scores $L_C=1$ against every
possible true class.

Both Convention A and Convention C confine their propriety defect to a
measure-zero set, but of different objects. Convention A's defect
(Proposition~\ref{prop:convention-a-failure}) is triggered by the \textit{forecast} landing exactly on a tie, for
any $Q$ — something a trained network's real-valued output almost never
does. Convention C's defect is triggered by $Q$ itself, the ground truth,
having a tied mode: a property of the data-generating process, not of a
forecaster's output, that does not shrink away with model precision or
training time. The two are not symmetric near-misses.

\begin{lemma}[Convention C attainment]
\label{lem:convention-c-attainment}
\textit{For both $S_{BS}$ and $S_{LL}$,
$\Sigma_\psi$ and $\Sigma_\xi$ take the same values under Convention C as
under Conventions A and B. Under Convention C, $\Sigma_\psi$ is never
attained and $\Sigma_\xi$ is always attained, for every $\omega=i$ — with no
order or distinguished class needed, unlike Convention B.}
\end{lemma}

\textit{Proof.} $\psi_C(i)\subseteq\psi_A(i)$, so $\sup_{\psi_C}\le\Sigma_\psi$. The
sequence used in Propositions~\ref{prop:sup-psi-brier} and~\ref{prop:sup-psi-log} to show approach-without-attainment
under Convention B for $i\ne j_0$ — one coordinate at $\frac1c+\varepsilon$,
every other strictly below it — has $M(p^{(\varepsilon)})=\{i\}$ for every
$\varepsilon>0$, hence lies in $\psi_C(i)$ too, and its score tends to
$\Sigma_\psi$ as $\varepsilon\to0^+$; so $\sup_{\psi_C}=\Sigma_\psi$. The
unique maximiser attaining $\Sigma_\psi$ under Convention A is the uniform
vector, a $c$-way tie, so it lies outside $\psi_C(i)$ for every $i$: the
supremum is never attained.

For the infimum, the extremal configuration of Propositions~\ref{prop:inf-xi-brier} and~\ref{prop:inf-xi-log} — one
other class tied with $i$ at the boundary value, every remaining coordinate
zero — has $M(p)=\{i,k\}\ne\{i\}$ for the tied class $k$, so it lies in
$\xi_C(i)$ by definition, for every $i$. Any $p$ with $M(p)\ne\{i\}$ either
excludes $i$ from the maximum outright or ties it there, and both cases give
some $k\ne i$ with $p_k\ge p_i$, so $\xi_C(i)\subseteq\overline{\xi_A(i)}$;
the lower bound $\Sigma_\xi$ established over $\overline{\xi_A}$ in
Propositions~\ref{prop:inf-xi-brier} and~\ref{prop:inf-xi-log} therefore applies to $\xi_C(i)$ directly, and since the
extremal configuration itself lies in $\xi_C(i)$ and attains that bound,
$\inf_{\xi_C}=\Sigma_\xi$, attained. $\square$

This attainment pattern is the reverse of Convention A's ($\Sigma_\psi$
attained, $\Sigma_\xi$ not) rather than "the same," and simpler than
Convention B's ($j_0$-dependent): under Convention C, $\Sigma_\xi$ is always
the attained extremum, uniformly in $i$. Since exactly one of the two
extrema is attained for every $\omega$, Proposition~\ref{prop:characterisation}(iii) gives superiority
at $\lambda=\lambda^\ast$ under Convention C directly — exactly the claim
Theorem~\ref{thm:minimality} below states for that convention.

\subsection{Minimality}
\label{sub:theory-2-minimality}

\begin{theorem}
\label{thm:minimality}
\textit{$\{\lambda\ge0:S_\lambda\text{ is superior}\}$ is an up-set
whose infimum is $\lambda^\ast:=\Sigma_\psi-\Sigma_\xi$, under Convention A or
Convention B. Whether $\lambda^\ast$ itself belongs to the set is
convention-dependent: it does under Convention A; it does not under
Convention B. Convention C also admits $\lambda^\ast$, by Lemma~\ref{lem:convention-c-attainment} above (the
proof below establishes Conventions A and B directly; the Convention C case
follows the same way once Lemma~\ref{lem:convention-c-attainment} supplies its attainment pattern).}
\end{theorem}

\textit{Proof.} Since the penalty is constant on $\xi$ and zero on $\psi$,
$\sup_\psi S_\lambda=\sup_\psi S_{\text{base}}=\Sigma_\psi$ and
$\inf_\xi S_\lambda=\inf_\xi S_{\text{base}}+\lambda=\Sigma_\xi+\lambda$, and
adding a constant does not change which extremum is attained. Applying
Proposition~\ref{prop:characterisation} (Section~\ref{sec:theory-1}) to $S_\lambda$: superiority holds iff
$\Sigma_\psi<\Sigma_\xi+\lambda$, or $\Sigma_\psi=\Sigma_\xi+\lambda$ with at
least one extremum unattained. For $\lambda>\lambda^\ast$ the inequality is
strict and $S_\lambda$ is superior by Proposition~\ref{prop:characterisation}(i). For $\lambda<
\lambda^\ast$, $\Sigma_\psi>\Sigma_\xi+\lambda$ and $S_\lambda$ is not superior
by Proposition~\ref{prop:characterisation}(ii). At $\lambda=\lambda^\ast$, equality holds, and Lemma~\ref{lem:attainment-dichotomy}'s
dichotomy is stated for a \textit{fixed} true class $\omega$; superiority instead
needs attainment across the \textit{union} over all $\omega$ (Proposition~\ref{prop:characterisation} again).
Under Convention A the extremal configuration — the uniform vector for
$\Sigma_\psi$, the pairwise tie for $\Sigma_\xi$ — sits on the same side of
the convention for every $\omega$, so the union attains only one extremum and
Proposition~\ref{prop:characterisation}(iii) gives superiority. Under Convention B, an order-induced
$\delta$ assigns the uniform vector to $\psi$ at $\omega=j_0$ (the
$\prec$-least class) but to $\xi$ at every other $\omega$, and symmetrically
for the pairwise tie; since $c\ge2$ guarantees some $\omega\ne j_0$, the union
attains $\Sigma_\psi$ at $\omega=j_0$ \textit{and} $\Sigma_\xi$ at $\omega\ne j_0$
simultaneously, so Proposition~\ref{prop:characterisation}(iii) gives \textbf{not superior}. $\square$

We call $\lambda^\ast$ the \textit{infimal penalty} throughout this paper, the
mathematically precise name for the quantity $\Sigma_\psi-\Sigma_\xi$ as a
greatest lower bound, and reserve \textit{minimal} or \textit{minimum} for the narrower
claim that this bound is actually attained. Theorem~\ref{thm:minimality} above gives
$\lambda^\ast$'s precise status by convention. It is a true minimum —
attained, and hence literally the smallest admissible penalty — under
Convention A. Under the recommended Convention B it is not attained: the
admissible set is the open interval $(\lambda^\ast,\infty)$, which has no
smallest element, and $\lambda^\ast$ is there only in the sense of being its
infimum. Superiority itself, under Convention B, therefore requires
$\lambda>\lambda^\ast$ strictly; $\lambda=\lambda^\ast$ is not admissible.
We do not call $\lambda^\ast$ itself the minimal admissible penalty except
where Convention A is explicitly in force, and we say so explicitly
wherever the distinction matters, as in the recommendation below.


\begin{corollary}[no choice of order rescues the boundary]
\label{cor:order-independence}
\textit{The failure of
$\lambda=\lambda^\ast$ under Convention B in Theorem~\ref{thm:minimality} does not depend on which
total order induces $\delta$: for every order $\prec$ and every $c\ge2$,
$S_{\lambda^\ast}$ is not superior under the resulting $\delta_\prec$.}
\end{corollary}

\textit{Proof.} Let $j_0$ be the $\prec$-least class. The uniform vector $u$ has
$\delta_\prec(u)=j_0$, so $(u,j_0)\in\psi$ with $S_{\lambda^\ast}(u,j_0)=
\Sigma_\psi$. For any two distinct classes $a,b$, let $v$ have
$v_a=v_b=\frac12$, other coordinates zero; then $\delta_\prec(v)$ is whichever
of $a,b$ is $\prec$-least, and taking $\omega$ to be the other gives
$(v,\omega)\in\xi$ with $S_{\lambda^\ast}(v,\omega)=\frac12+\lambda^\ast=
\Sigma_\xi+\lambda^\ast=\Sigma_\psi$ by direct computation. The pair $(u,j_0)$,
$(v,\omega)$ has equal scores, violating the strict inequality the source
paper's own Definition 5 requires, for every choice of $\prec$. $\square$

Numerically, the margin $\min_\xi S_{\lambda^\ast}-\max_\psi S_{\lambda^\ast}$ is
exactly zero to floating-point precision under four different orders
(lowest-index, highest-index, a middle-first order, and a randomly generated
order) at $c\in\{3,5,8,13\}$, and never positive.

\begin{corollary}[the admissible set under the recommended convention]
\label{cor:admissible-set}
\textit{Under
Convention B, $\{\lambda:S_\lambda\text{ superior}\}=(\lambda^\ast,\infty)$, an
open interval.}
\end{corollary}

\begin{corollary}[closed forms]
\label{cor:closed-forms}
\begin{equation}\lambda^\ast_{PBS}=\frac{c-1}{c}-\frac12=\frac{c-2}{2c},\end{equation}
\begin{equation}\lambda^\ast_{PLL}=\log c-\log2=\log\frac{c}{2}\ \ (p\in\Delta_c^\circ).\end{equation}
Both vanish at $c=2$ (Corollary~\ref{cor:c2}, theory, part 1). As $c\to\infty$,
$\lambda^\ast_{PBS}\to\frac12$, while the published $\frac{c-1}{c}\to1$.
\end{corollary}

\begin{corollary}[positioning the published rules]
\label{cor:positioning-published}
\textit{$PBS$ and $PLL$ are the
members of the family at $\lambda=\Sigma_\psi$, i.e. the penalty set from the
supremum alone with $\Sigma_\xi$ taken implicitly as zero ($PLL$'s claim is
for $p\in\Delta_c^\circ$, preliminaries). Since
$\Sigma_\psi>\lambda^\ast$ whenever $\Sigma_\xi>0$, they lie strictly inside the
admissible set of Corollary~\ref{cor:admissible-set} and are therefore superior under every
convention, matching the source paper's own claim (source Theorem 6) that
these penalties are superior, but not minimal.}
\end{corollary}

We state $\lambda^\ast$ throughout as an infimum of admissible penalties, never
asserting it is attained without naming a convention, since Theorem~\ref{thm:minimality} shows
attainment genuinely depends on which one is in force.


\section{Relation to the published rules}
\label{sec:relation-to-published}

\subsection{Positioning $PBS$ and $PLL$ within the family}
\label{sub:relation-to-published-positioning-pbs-and-pll-within-the-family}

Corollary~\ref{cor:positioning-published} (Section~\ref{sec:theory-2}) places the published rules precisely within the
family introduced here: $PBS$ and $PLL$ are the members of $S_\lambda$ at
$\lambda=\Sigma_\psi$, the penalty set from the supremum of the base rule
over $\psi$ alone, with $\Sigma_\xi$ taken implicitly as zero. Since
$\Sigma_\xi>0$ in both cases considered here, $\lambda=\Sigma_\psi$ is
strictly greater than $\lambda^\ast=\Sigma_\psi-\Sigma_\xi$, so it lies inside
the admissible set under either tie convention (Theorem~\ref{thm:minimality}, theory, part 2) —
the boundary subtlety that convention alone decides does not touch the
published rules. $PBS$ and $PLL$ are therefore superior, matching the
source paper's own claim (source Theorem 6), and our framework identifies exactly why: the published
construction is a special case of a one-parameter family, calibrated from
$\Sigma_\psi$ alone and landing strictly above the infimum $\lambda^\ast$
rather than at it. The admissible set has no upper bound
($\{\lambda:S_\lambda\text{ superior}\}=(\lambda^\ast,\infty)$ under the
recommended convention, Corollary~\ref{cor:admissible-set}), so there is no largest admissible
penalty for the published construction to be; it is simply one specific,
finite point well inside an unbounded admissible range, not a fixed design
choice tied to any upper limit.

\begin{table}[htbp]
\centering
\caption{The calibration facts of this section, gathered in one place, at $c=13$ (the value used throughout Figures 1, 2, and 4).}
\label{tab:relation-to-published-1}
\begin{tabularx}{\textwidth}{YYY}
\hline
 & Brier ($PBS$) & Log ($PLL$) \\
\hline
Source def. & Eq. 42, $\lambda=\Sigma_\psi=\frac{c-1}{c}$ & Eq. 43, $\lambda=\Sigma_\psi=\log c$ \\
Published $\lambda$ & $0.923$ & $2.565$ \\
Infimal $\lambda^\ast$ & $0.423$ & $1.872$ \\
Ratio, published/$\lambda^\ast$ & $2.18\times$ & not tabulated \\
Implementation & Algorithm 2 & Algorithm 3 \\
Impl. vs. published & Mean-form; same $\lambda$ & Matches published exactly \\
Sum-form-equiv. penalty & $c-1=12.000$ & n/a (no mean/sum split) \\
Ratio, implementation/$\lambda^\ast$ & $28.36\times$ & — \\
Role in this paper & Member of $S_\lambda$, above $\lambda^\ast$ & Same \\
\hline
\end{tabularx}
\end{table}

Table~\ref{tab:relation-to-published-1} collects every fact this section establishes in the prose that
follows.

\subsection{The gap between the published penalty and $\lambda^\ast$}
\label{sub:relation-to-published-the-gap-between-the-published-penalty-and-lambda}

Because the published penalty is set from $\Sigma_\psi$ alone, without
reference to $\Sigma_\xi$, it exceeds the infimal penalty $\lambda^\ast$ by a
margin that widens as $c$ shrinks. For the Brier base, the ratio of the
source paper's Eq.\ 42 penalty to $\lambda^\ast$ runs from $4.00$ at $c=3$ down to $2.11$
at $c=20$ (at $c=13$, $2.18$); we return below to the released
implementation's effective penalty, whose ratio to $\lambda^\ast$ is
substantially larger again. The infimum $\Sigma_\xi$ was simply not part of
the published construction; computing it, as we do here, recovers a tighter
constant while leaving the superiority guarantee intact.

\begin{figure}[htbp]
\centering
\includegraphics[width=3.3in]{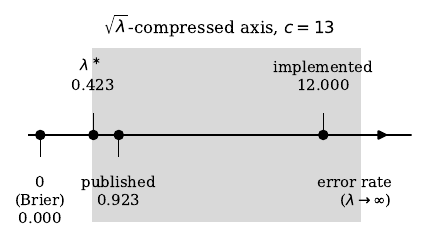}
\caption{The admissible range for $\lambda$ at $c=13$: zero, the infimal
penalty $\lambda^\ast=0.423$, the published penalty $0.923$, and the
released implementation's sum-form-equivalent penalty $12.000$. The axis
position of each value is $\sqrt{\lambda}$, not $\lambda$ itself,
compressing the wide gap
to the implemented penalty for display; axis labels give the true values.
The superiority region $\lambda>\lambda^\ast$ is shaded, and the
error-rate limit $\lambda\to\infty$ (equivalent to monitoring error rate
directly) is marked off the right edge rather than placed at a finite
position.}
\label{fig:lambda-dial}
\end{figure}

Figure~\ref{fig:lambda-dial} places all three penalty values on one axis at
$c=13$: the published penalty already clears $\lambda^\ast$ by a wide margin,
and the implemented penalty clears it by far more.

\subsection{Sum form and mean form}
\label{sub:relation-to-published-sum-form-and-mean-form}

Four distinct quantities recur in this section and should not be read as
interchangeable: the base's theoretical normalisation (sum-form throughout
this paper, per preliminaries); the published penalty, the number Eq.\ 42
states for a given $c$; the implemented penalty, the number Algorithm 2's
released code actually pairs with its own (mean-form) base; and the
effective numerical penalty, what that pairing amounts to once converted
to the same sum-form units as $\lambda^\ast$. The published and implemented
penalties turn out to share one number here — Algorithm 2 pairs the
published value with a different base, not a different number — but the
effective penalty is a third, larger number again, and it is the effective
penalty, not the published one, that belongs on the same axis as
$\lambda^\ast$.

Source Eq.\ 42 specifies the penalty against sum-form Brier.
Algorithm 2, which the released implementation follows, pairs the same
penalty value with a mean-form base. Since the mean form is smaller than the
sum form by a factor of $c$, pairing it with a sum-form-calibrated penalty is
equivalent, in sum-form terms, to a penalty of $c-1$ rather than
$\frac{c-1}{c}$: at $c=3$ this is $2$, and at $c=13$ it is $12$. This
sum-form-equivalent penalty exceeds the entire range of the base calibration
term itself — the width of $\psi$'s range is at most $\Sigma_\psi=\frac{c-1}{c}<1$
— by a factor of $c$; at $c=13$, the penalty is thirteen times the width of
the calibration term it is meant to supplement. All experiments reported in
the source paper used this implemented form. We state this as a fact with a
direct interpretive consequence: under the implemented penalty, a scoring
rule intended to combine calibration and decision quality orders models
essentially by error count, since the calibration term contributes at most a
fraction $1/c$ of what the decision term contributes. This affects only the
balance between the two terms, not the guarantee itself: the implemented
sum-form-equivalent penalty $c-1$ is well clear of $\lambda^\ast$ under either
tie convention (Theorem~\ref{thm:minimality}, theory, part 2), so the reported experiments
remain a valid instance of a superior rule — the open question this raises is
which value along the admissible range was actually being measured. The
logarithmic base has no analogous gap: Algorithm 3 computes
$-\log(q_\omega)+\log(c)\cdot\mathbb{1}[\text{misclassified}]$ exactly,
matching Eq.\ 43's own construction, since it subtracts the negative
constant $\log(1/c)$ rather than adding a positive one — algebraically
identical, with no mean/sum-form split to introduce a gap the way Algorithm
2 does. Table~\ref{tab:relation-to-published-2} sets the Brier-base gap out across every class count
already used elsewhere in this paper.

\begin{table}[htbp]
\centering
\caption{How far each Brier-base penalty sits from $\lambda^\ast$, across the class counts already used elsewhere in this paper.}
\label{tab:relation-to-published-2}
\begin{tabular}{lccccc}
\hline
$c$ & $\lambda^\ast$ & Published $\lambda$ & Ratio & Implemented $\lambda$ & Ratio \\
\hline
$3$ & $0.167$ & $0.667$ & $4.00\times$ & $2.0$ & $12.00\times$ \\
$10$ & $0.400$ & $0.900$ & $2.25\times$ & $9.0$ & $22.50\times$ \\
$13$ & $0.423$ & $0.923$ & $2.18\times$ & $12.0$ & $28.36\times$ \\
$20$ & $0.450$ & $0.950$ & $2.11\times$ & $19.0$ & $42.22\times$ \\
$47$ & $0.479$ & $0.979$ & $2.04\times$ & $46.0$ & $96.09\times$ \\
\hline
\end{tabular}
\end{table}

Both ratios shrink as $c$ grows, but the implemented penalty's ratio stays
an order of magnitude above the published penalty's own at every $c$ shown
— the mean/sum-form gap, not the choice of calibration point, dominates the
implementation's distance from $\lambda^\ast$.

\subsection{The proof of source Theorem 5}
\label{sub:relation-to-published-the-proof-of-source-theorem-5}

Source Theorem 5 states that $PBS$ and $PLL$ are strictly proper. The theorem
is true; Theorem~\ref{thm:strict-propriety} (Section~\ref{sec:theory-2}) establishes it for the entire family
$S_\lambda$, of which $PBS$ and $PLL$ are two instances, so nothing here
revises the published claim. The published argument rests on the identity
$S_{BS}(Q,Q)=S_{PBS}(Q,Q)$ (source Eq. 46). This identity does not hold in
general: a truthful forecaster still incurs the penalty on every realisation
whose class is not the mode of $Q$, an event of probability $1-\max_jQ_j$,
positive unless $Q$ is degenerate. The corrected identity is
\begin{equation}S_{PBS}(Q,Q)=S_{BS}(Q,Q)+\lambda\big(1-\max_jQ_j\big),\end{equation}
which follows directly from expanding the expectation:
$S_{PBS}(Q,Q)=\sum_iQ_i\big[S_{BS}(Q,i)+\lambda\mathbb{1}[i\ne\delta(Q)]\big]
=S_{BS}(Q,Q)+\lambda(1-Q_{\delta(Q)})$, and $Q_{\delta(Q)}=\max_jQ_j$. At
$c=3$, $Q=(0.5,0.3,0.2)$, the two sides of source Eq. (46) differ by
$0.333$, matching $\lambda(1-\max_jQ_j)$ exactly; equality would require
$\max_jQ_j=1$, a degenerate forecast, the one case in which probabilistic
evaluation is uninteresting. We supply this corrected identity in place of
the identity the published proof relies on.

\subsection{Two further transcription matters}
\label{sub:relation-to-published-two-further-transcription-matters}

Two further matters are minor enough to note together rather than give
each its own point. Source
Eq. (50) compares a quantity with itself, and Eq. (51) carries the Brier
notation into the logarithmic-loss chain; both read as transcription slips
rather than substantive claims. Separately, source Table~\ref{tab:preliminaries-1}'s log-loss
values are computed in base 10 rather than the natural log used in the
penalty and in
the rest of this paper ($-\log_{10}(0.34)=0.469$, matching the published
value, against $-\ln(0.34)=1.079$); both source Table~\ref{tab:preliminaries-1}'s Brier values
reproduce exactly in sum form, confirming the source paper's Eq.\ 42
as the intended object. These are courtesies to
a future reader rather than arguments, and we record them as such.


\section{Experimental design}
\label{sec:experimental-design}

We pre-specify the design below before reading any result. A formal
pre-registration freeze was intended but did not literally occur. 

\subsection{The penalty grid}
\label{sub:experimental-design-the-penalty-grid}

We sweep $\lambda$ across a small grid, anchored at three points the
theory sections already establish: zero, the infimal penalty $\lambda^\ast$,
and the published value. We add a fourth anchor, the exact penalty
magnitude the source paper's own experiments ran under (labelled
implemented), since a practitioner comparing against the source paper's
reported results should see the frontier measured at that specific value
too, not only at the value it states. We also adopt the single-valued tie
convention throughout, since it is the only one Theorem~\ref{thm:strict-propriety} (Section~\ref{sec:theory-2})
makes strictly proper for every $\lambda$ (Propositions~\ref{prop:convention-a-failure} and~\ref{prop:convention-c-failure} rule the
other two out). Table~\ref{tab:experimental-design-1} gives every grid point's value for $c=10$, the
class count both datasets share.

\begin{table}[htbp]
\centering
\caption{The $\lambda$ grid at $c=10$. Supercritical sits immediately above $\lambda^\ast$, since Theorem~\ref{thm:minimality} (Section~\ref{sec:theory-2}) makes $\lambda^\ast$ itself convention-dependent for attainment, and testing the boundary needs a point unambiguously inside the admissible set under every convention.}
\label{tab:experimental-design-1}
\begin{tabularx}{\textwidth}{>{\raggedright\arraybackslash}p{1.44in}>{\centering\arraybackslash}p{0.93in}>{\centering\arraybackslash}p{0.68in}Y}
\hline
Grid point & Brier $\lambda$ & Log $\lambda$ & What it is \\
\hline
zero & $0$ & $0$ & No penalty; the base rule alone \\
critical ($\lambda^\ast$) & $0.400$ & $1.609$ & The infimal penalty (Corollary~\ref{cor:closed-forms}, theory part 2) \\
supercritical & $0.400001$ & — & $\lambda^\ast+10^{-6}$: strictly admissible \\
published & $0.900$ & $2.303$ & The source paper's own calibration \\
implemented & $9.000$ & n/a & The exact magnitude the source paper's own experiments ran under, on the Brier base; the log base has no comparable value to report (relation to published rules) \\
limit & $50.0$ & $50.0$ & Approximates $\lambda\to\infty$, i.e. monitoring error rate directly \\
\hline
\end{tabularx}
\end{table}

\subsection{Datasets and regimes}
\label{sub:experimental-design-datasets-and-regimes}

The design uses MNIST and CIFAR-10, both with $c=10$. An earlier
nine-dataset design matched the source paper's own class counts, and we
record narrowing it to two as a deliberate deviation, not an omission.
This means the study does not test the theory's sharpest quantitative
prediction — the published-to-$\lambda^\ast$ ratio falls from $4.00$ at
$c=3$ to $2.11$ at $c=20$ (Corollary~\ref{cor:closed-forms}, theory part 2) — since that
prediction is a closed form, not something any dataset could test. A
third dataset, with a different number of classes, was added after the
two-dataset design had been finalized. We therefore document its
inclusion as a dated and reasoned amendment to the original design,
consistent with the transparency principle used in pre-registered
clinical and psychological studies to disclose post-hoc protocol changes.
We refer to this change throughout the paper as Amendment 1.

At full training strength both datasets already exceed $90\%$ macro-F1,
where misclassifications and near-ties are too rare for the penalty term
to matter. So each dataset is also run under three harder regimes, giving
four per dataset in total (Table~\ref{tab:experimental-design-2}). A regime whose achieved macro-F1
still exceeds $90\%$ is reported but excluded from the headline analysis,
since it cannot inform the frontier either way; this threshold was fixed
before any result was read.

\begin{table}[htbp]
\centering
\caption{The four regimes, crossed with the $\lambda$ grid for each dataset.}
\label{tab:experimental-design-2}
\begin{tabularx}{\textwidth}{>{\raggedright\arraybackslash}p{0.76in}YY}
\hline
Regime & What changes & Why \\
\hline
full & Nothing — full training data, full network capacity & The dataset's natural difficulty \\
subset & Training data reduced & Fewer examples, more room for near-ties \\
noisy & Symmetric label noise injected & A different, independent way to create near-ties \\
hard & Reduced data and label noise together & Both pressures combined \\
\hline
\end{tabularx}
\end{table}

The subset, noisy, and hard regimes also use a deliberately under-capacity
network, so predictions stay diffuse rather than sharpening back toward
full confidence.

\subsection{What is held fixed}
\label{sub:experimental-design-what-is-held-fixed}

Every model is trained once per dataset, regime, and seed, using categorical
cross-entropy as the training loss; $\lambda$ never enters training; because
the penalty term has zero gradient almost everywhere, it functions here only
as an evaluation and model-selection criterion. Per-epoch validation
predictions are cached, and every point on the $\lambda$ grid is scored
offline against the same cached stream, so $\lambda$ is the only factor that
varies between selections drawn from a given training run. Early stopping is
likewise simulated offline from the cached stream rather than run live for
each $\lambda$ separately. Optimiser, learning rate, batch size, and
initialisation are fixed within each (dataset, regime) cell and identical
across seeds; five seeds are trained per cell and all five are reported,
none dropped. Cross-validation uses stratified holdout rather than the
source paper's h-block scheme, a deliberate deviation recorded here rather
than treated as an oversight: MNIST and CIFAR-10 are i.i.d. image datasets
with no temporal dependence, so h-block cross-validation is unnecessary for
them.

\subsection{Outcome measures}
\label{sub:experimental-design-outcome-measures}

Four measures are reported for the model selected under each $\lambda$:
selected-model macro-F1, expected calibration error, test negative
log-likelihood, and the seed-to-seed variance of the selected epoch. The
source paper reports only the first of these. We use macro-F1, not raw
accuracy, as our operationalisation of the target decision utility —
classification correctness, not calibration — because several regimes are
constructed to be class-imbalanced (the noisy and reduced-data regimes
among them), where macro-F1 is the more informative of the two. We name it
consistently by this name throughout the rest of the paper rather than
call it accuracy, since the two are not the same statistic and macro-F1's
own class-balancing is part of what several regimes are designed to
exercise. Reporting all four is what makes a genuine
calibration-versus-macro-F1 trade-off visible, if one exists,
rather than collapsing the question to whether macro-F1 alone moves. The
fourth measure addresses stability specifically: a monitoring metric that
changes discontinuously at an argmax boundary can select a different epoch
whenever a single validation sample crosses that boundary, and this measure
is designed to detect that. Selection itself uses validation performance,
as throughout; all four outcome measures are then computed on the held-out
test split for whichever checkpoint that selection landed on, so they
report generalisation of the selected model rather than the validation
signal that chose it. Selection is run under two modes, checkpointing and
early stopping, matching the source paper's own split, and both are
reported separately rather than pooled. Every reported difference is
accompanied by its dispersion across seeds; no difference is described as an
improvement without it.

Expected calibration error is confidence-based, not classwise: for each
test example we take the predicted class's own probability,
$\max_j p_j$, as its confidence, bin examples into $15$ equal-mass bins
by sorting on this confidence (so each bin holds an equal count of
examples rather than an equal width of the $[0,1]$ probability range, and
no bin is ever empty by construction), and within each bin compare mean
confidence against the bin's observed top-1 accuracy. The reported value is the standard weighted-average gap,
$\mathrm{ECE}=\sum_{\text{bins}}\frac{n_{\text{bin}}}{n}\,
\lvert\text{mean confidence}-\text{accuracy}\rvert$. This is our own
construction, fixed in the design before any result was read; the source
paper reports no calibration measure of its own, proposing histogram-based
measures such as ECE only as future work (source Section 6.6; discussion).

\subsection{Baselines}
\label{sub:experimental-design-baselines}

Three baselines accompany the $\lambda$-grid comparison (Table~\ref{tab:experimental-design-3}). The
third was meant to be the design's most important one: without it, a
frontier that tracks $\lambda$ cannot distinguish two explanations for why
— that $S_\lambda$ approximates macro-F1 by construction as $\lambda$
grows, or that the superiority property specifically is doing the work.
It does not succeed at this, for a reason this design did not anticipate
(Section~\ref{sub:results-baselines-does-the-frontier-require-superiority-or-only-re-weighting}; Section~\ref{sec:limitations}).

\begin{table}[htbp]
\centering
\caption{The three baselines.}
\label{tab:experimental-design-3}
\begin{tabularx}{\textwidth}{YYY}
\hline
Baseline & What it measures & Purpose \\
\hline
Monitor macro-F1 directly & Select the checkpoint with the best macro-F1, full stop & Upper bound on macro-F1-driven selection \\
Monitor error rate directly & Select on error rate alone & The $\lambda\to\infty$ limit of the family \\
Weighted mixture $M_w$ & Brier and error rate, blended at the same weights $S_\lambda$ sweeps, but without superiority & Meant to isolate superiority from re-weighting; turned out not to (Section~\ref{sub:results-baselines-does-the-frontier-require-superiority-or-only-re-weighting}) \\
\hline
\end{tabularx}
\end{table}

\subsection{Diagnostics for a null result}
\label{sub:experimental-design-diagnostics-for-a-null-result}

Because a flat frontier is one of the three pre-specified outcomes below, the
design also records, per regime and $\lambda$, the margin between the top two
predicted class probabilities, the fraction of validation samples within
$0.01$ of a tie, the fraction of samples on which the penalty actually fires,
and the achieved macro-F1 used to classify a regime as informative or not.
These distinguish a flat frontier that arises because near-ties are absent
from one that arises because the penalty genuinely does not affect selection
once $\lambda$ clears $\lambda^\ast$; without them, the two are indistinguishable
from the headline numbers alone.

\subsection{What this experiment can and cannot establish about superiority}
\label{sub:experimental-design-what-this-experiment-can-and-cannot-establish-about-superiority}

Superiority (Section~\ref{sec:theory-1}) is a pointwise ordering guarantee on individual
predictions; it says nothing, by itself, about how a class-averaged,
nonlinear aggregate like macro-F1 behaves as $\lambda$ varies. The
penalty term $\lambda\cdot\mathbb{1}[p\in\xi]$ is likewise an unweighted,
per-example misclassification indicator, not a macro-F1-aware quantity,
and no theorem in this paper connects the two. The outcomes below —
"superiority is the active mechanism" versus "superiority is incidental" —
are therefore empirical labels for patterns in macro-F1, not statements this
design derives from the theory; the empirical section that follows is
motivated by association, not theoretical consequence — we investigate
whether this penalty improves macro-F1 in the conditions tested here, and
the theory's own guarantees hold regardless of what this investigation
finds.

\subsection{Three outcomes, stated as equally acceptable before any result is read}
\label{sub:experimental-design-three-outcomes-stated-as-equally-acceptable-before-any-result-is-read}

We pre-specify three possible outcomes for how macro-F1 behaves as
$\lambda$ grows past $\lambda^\ast$, and commit in advance to treating all
three as equally acceptable findings, not as a hypothesis we are trying to
confirm (Table~\ref{tab:experimental-design-4}).

\begin{table}[htbp]
\centering
\caption{The three pre-specified outcomes.}
\label{tab:experimental-design-4}
\begin{tabularx}{\textwidth}{>{\raggedright\arraybackslash}p{0.85in}Y>{\centering\arraybackslash}p{1.87in}Y}
\hline
Outcome & Macro-F1 past $\lambda^\ast$ & Calibration and NLL & What it would mean \\
\hline
flat & Statistically flat & Still degrade & Superiority is the active mechanism; the published penalty spends calibration sensitivity for no macro-F1 gain \\
rising & Keeps rising, even past published & Still degrade & Superiority is incidental to whatever benefit the penalty gives; monitor error rate or macro-F1 directly instead \\
peaked & Peaks at an interior $\lambda$ & Still degrade & A genuine trade-off; the contribution is a $\lambda$-selection rule, not one recommended value \\
\hline
\end{tabularx}
\end{table}

The "what it would mean" column states the interpretation as it was
pre-specified; we leave the three outcome definitions themselves
unedited, as a pre-registration should be, rather than revise them in
light of what came later. But the causal-mechanism language in that
column — "superiority is the active mechanism," "superiority is
incidental" — presumes an instrument capable of isolating superiority
from the penalty's other effect, re-weighting toward misclassification.
The baseline meant to supply that instrument turned out not to (baselines,
below; results, baselines): none of the three outcomes, on its own,
licenses attributing what is observed to superiority specifically. We
report which outcome obtains, in results, without relying on this column
to interpret it.

We record in advance that flat is our own motivating hypothesis, so
rising obtaining is a negative result for it — and, if it obtains,
the more interesting finding, stated in the same register either way. The
primary comparison — macro-F1 at the point just above $\lambda^\ast$ against
macro-F1 at the published value, paired by dataset, regime, seed, and
selection mode — is tested by a Wilcoxon signed-rank test at
$\alpha=0.05$ with a Holm correction for multiple comparisons, fixed in
the design so the test cannot be chosen after the data are seen.

This pairing has a nesting structure worth stating explicitly, since it
determines the effective sample size: five seeds per (dataset, regime)
cell are independent training runs, but the two selection modes read off
the same trained run rather than replicate it, and both are nested within
just nine such cells (eight once emnist\_balanced, added afterward by
Amendment 1, is set aside as exploratory — results). The
(dataset, regime) cell, not the individual (seed, selection-mode) pair, is
therefore the effective confirmatory unit; the pairing above generates
$90$ observations but at most $8$ independent ones. We designed the test
this way before recognising the consequence, and report both the
pair-level and cell-level analyses in results for exactly this reason,
with the cell-level one treated as the sole inferential basis for this
paper's conclusions.

\subsection{Scope}
\label{sub:experimental-design-scope}

This design departs from the source paper's own evaluation in two respects
that bear on how its results should be read against the source paper's: it
evaluates on images rather than the source paper's spatio-temporal sensor
data, so transfer of any conclusion is an assumption rather than a
demonstration; and its label-noise regime produces near-ties by a different
mechanism than genuine class overlap, so the regime axis may not reproduce
the tie structure of the source paper's own datasets. Both points are
stated here because they bound what the experiment can show, being
properties of the design itself rather than of any result it produces.


\section{Results}
\label{sec:results}

This section reports the paper's tertiary contribution: a pre-specified,
bounded case study testing what the theory above implies in practice, not
a further theoretical claim, and not a premise the paper's admissibility
or propriety results depend on — those hold regardless of anything
reported below. Every number in this section is derived from the frozen
experimental records by a single fixed analysis procedure, with no manual
computation; this mirrors how the figures are derived directly from the
same closed forms established in the theory sections.

The design covers three datasets across two configuration versions: mnist
and cifar10, both pre-specified together and complete under one
configuration hash (40 cells); and emnist\_balanced, added by Amendment 1
specifically to test whether the class-count dependence the theory predicts
shows up in the measured frontier, complete under a second configuration
hash (5 cells, full regime only, 5 seeds). Our analysis checks
configuration-hash consistency per cell, not per file, since the two
versions legitimately coexist.

\subsection{Which regimes inform the frontier}
\label{sub:results-which-regimes-inform-the-frontier}

Of the nine (dataset, regime) combinations, four exceed the pre-specified
$90\%$ validation macro-F1 threshold and are excluded from the headline
analysis, consistent with the design's own prediction that near-ceiling
macro-F1 leaves the penalty term nothing to discriminate. The remaining
five are informative; emnist\_balanced/full's own informativeness confirms,
on its own, the design's separate prediction that it would not need a
reduced-data or noisy regime to become informative, unlike mnist and
cifar10 at full strength. Restricted to the five informative regimes, the
diagnostic measures confirm the regime axis is doing its job: near-tie
rate and penalty-fire rate are both close to zero on every excluded
regime and substantially higher on every included one, with no overlap
between the two groups. Table~\ref{tab:results-1} gives every cell; we report all nine
regimes throughout, but the hypothesis test below and the frontier
figures that follow describe the five informative ones unless stated
otherwise.

\begin{table}[htbp]
\centering
\caption{Every (dataset, regime) cell: mean validation macro-F1 (the informativeness criterion), whether it clears the $90\%$ threshold, the near-tie rate (fraction of validation samples within $0.01$ of a top-two tie), and the penalty-fire rate (fraction of validation samples on which the $\lambda$ penalty actually changes which class is credited).}
\label{tab:results-1}
\begin{tabular}{lcccc}
\hline
Dataset/regime & Macro-F1 & Informative & Near-tie rate & Penalty-fire rate \\
\hline
mnist/full & $0.9953$ & no & $0.0\%$ & $0.5\%$ \\
mnist/subset & $0.9818$ & no & $0.0\%$ & $1.8\%$ \\
mnist/noisy & $0.9815$ & no & $0.3\%$ & $1.8\%$ \\
mnist/hard & $0.8830$ & yes & $1.2\%$ & $11.6\%$ \\
cifar10/full & $0.9144$ & no & $0.1\%$ & $8.6\%$ \\
cifar10/subset & $0.6080$ & yes & $2.8\%$ & $38.9\%$ \\
cifar10/noisy & $0.6422$ & yes & $12.2\%$ & $34.8\%$ \\
cifar10/hard & $0.4152$ & yes & $18.7\%$ & $57.1\%$ \\
emnist\_balanced/full & $0.8922$ & yes & $0.4\%$ & $10.6\%$ \\
\hline
\end{tabular}
\end{table}

\subsection{Primary comparison: supercritical vs. published macro-F1 (cell-level analysis)}
\label{sub:results-primary-comparison-supercritical-vs-published-macro-f1-cell-level-analysis}

We compare supercritical against published macro-F1 with the (dataset, regime)
cell as the inferential unit, and treat this as the only analysis this
design licenses drawing an inferential conclusion from. Seeds are
independent training runs, but the two selection modes read off the same
trained run rather than replicate it, so pairing by (dataset, regime, seed,
mode) is nested within a handful of independent cells, not ninety; we
collapse each cell's ten (seed, mode) differences to one cell-level mean
difference before comparing. We further distinguish the confirmatory cells
from an exploratory one: the design frozen before any result was read
covered mnist and cifar10 only, four regimes each, $n=8$ cells; emnist\_balanced
was added afterward by Amendment 1, and pooling its one cell into the
"primary" statistic would conflate confirmatory and exploratory evidence.
We report the $n=8$ confirmatory analysis as primary and the full $n=9$
(with emnist\_balanced) as a check on whether the amendment changes the
conclusion, not as a second, equally-weighted estimate. Even $n=8$
overstates independence in one further respect worth stating plainly: the
eight cells are four regimes each on only two underlying datasets, not
eight draws from the broader problem domain. A regime is a manipulation
of an existing dataset (a reduced-data subset, a noise injection), not a
new sample of "classification tasks in general," so mnist's four cells
share whatever is idiosyncratic to mnist, and likewise for cifar10. We
report cell-level statistics because the cell is the finest unit this
design's pairing structure supports, not because we take $n=8$ to mean
eight independent replicates of the underlying question; the safe reading
of everything below is as a case study on two datasets, not a
population-level estimate of a general effect.

The confirmatory ($n=8$) effect estimate is small and its interval
straddles zero for both bases. For the Brier base, the mean cell-level
difference (supercritical minus published) is $-0.0036$ (median $0.0000$,
standard deviation $0.0067$), $95\%$ $t$-interval $[-0.0092, 0.0020]$ and
bootstrap interval $[-0.0090, 0.0000]$ — both include zero, or in the
bootstrap case touch it exactly at the boundary. For the logarithmic base,
the mean difference is $-0.0034$ (median $-0.0001$, standard deviation
$0.0054$), $t$-interval $[-0.0078, 0.0011]$ (includes zero) and bootstrap
interval $[-0.0071, -0.0005]$ (does not). Table~\ref{tab:results-2} gives the full picture; a
Wilcoxon signed-rank test on the eight cell-level differences, kept as a
secondary diagnostic, agrees (Brier $W=1.0$, $p=0.50$; log $W=1.0$,
$p=0.125$, Holm-adjusted $p=0.25$), as does an exact sign test over the
non-tied cells (Brier $p=1.0$; log $p=0.375$) — of the eight confirmatory
cells, only one favours supercritical for either base (two favour published
for Brier, four for log; the rest are exactly tied). Neither base reaches
$\alpha=0.05$.

Including emnist\_balanced ($n=9$) does not change this conclusion: the
mean difference shifts slightly toward zero for both bases ($-0.0031$
Brier, $-0.0029$ log, since emnist\_balanced's own cell favours supercritical
slightly), and both intervals and both tests remain non-significant (Brier
$W=3.0$, $p=0.625$; log $W=4.0$, $p=0.219$, Holm-adjusted $p=0.438$;
sign test Brier $p=1.0$, log $p=0.688$). We report this as a stability
check on the substantive conclusion, not as additional confirmatory
evidence — the amendment cell remains exploratory regardless of which way
it happens to point.

We treat the bootstrap interval throughout as a sensitivity check on the
$t$-interval, not as an equally weighted second estimate: with a handful
of higher-level units, a percentile bootstrap resamples from very few
values, and its endpoints are correspondingly unstable — the log base's
own disagreement between the two intervals, at both $n=8$ and $n=9$, is
itself evidence of that instability, not a more precise finding than the
$t$-interval gives. Taken together, this evidence does not support a
robust inferential conclusion in either direction at the cell level, at
either $n$; we do not treat the log base's bootstrap interval excluding
zero as outweighing the $t$-interval, the Wilcoxon test, and the sign test
all failing to reject. Nor was this design formally powered at $n=8$: no
power analysis was run before the sweep, and with eight independent units
we cannot say with confidence that the design could detect a true effect
of the size the observed means suggest, let alone a smaller one. Failing
to reject at this $n$ is therefore not itself evidence for a true null
effect; it is one further respect in which this design supports
description over confirmation.

\begin{table}[htbp]
\centering
\caption{Cell-level effect of supercritical versus published on macro-F1, confirmatory $n=8$ (dataset, regime) cells (mnist and cifar10 only; see text for the $n=9$ check with emnist\_balanced included), each a mean over that cell's five seeds and two selection modes. Cells column is favour-supercritical/favour-published/tied; Wilcoxon and sign-test $p$-values (prose above) are secondary diagnostics, not shown here.}
\label{tab:results-2}
\begin{tabular}{lcccccc}
\hline
Base & Mean diff. & Median & SD & 95\% CI ($t$) & 95\% CI (bootstrap) & Cells \\
\hline
Brier & $-0.0036$ & $0.0000$ & $0.0067$ & $[-0.0092, 0.0020]$ & $[-0.0090, 0.0000]$ & $1/2/5$ \\
Logarithmic & $-0.0034$ & $-0.0001$ & $0.0054$ & $[-0.0078, 0.0011]$ & $[-0.0071, -0.0005]$ & $1/4/3$ \\
\hline
\end{tabular}
\end{table}

\begin{figure}[htbp]
\centering
\includegraphics[width=5.2in]{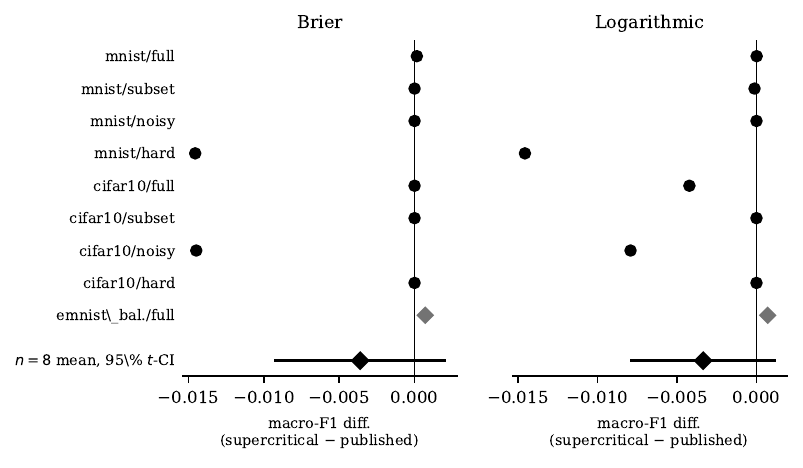}
\caption{Cell-level macro-F1 difference (supercritical minus published)
for each of the nine (dataset, regime) cells, both bases. Circles are the
eight confirmatory cells (mnist and cifar10, four regimes each); the
diamond is the exploratory emnist\_balanced/full cell, added afterward by
Amendment 1 and not pooled into the primary statistic. The bottom row
shows the confirmatory ($n=8$) mean difference with its 95\% $t$-interval,
the same numbers Table 10 reports. Both intervals straddle zero; most
individual cells are exactly tied (difference $0$) or favour published by
a small margin.}
\label{fig:primary-comparison-cells}
\end{figure}

Figure~\ref{fig:primary-comparison-cells} shows this cell by cell rather
than only in aggregate: most cells sit exactly at zero (the tied checkpoint
selections behind the counts in Table~\ref{tab:results-2}'s rightmost column), the two
cells that do move both favour published, and the exploratory
emnist\_balanced/full cell (diamond) sits on the other side of zero from
the confirmatory mean — one more reason its own $n=1$ status keeps it out
of the primary statistic.

Two informative $c=10$ regimes (cifar10/hard and cifar10/subset, Brier
base) are exactly tied at the cell level alongside the non-informative
ones, so the ties are not confined to regimes already known to be
uninformative. We do not claim a statistically robust improvement
associated with increasing $\lambda$ past $\lambda^\ast$: at this unit, the
effect is small, its interval straddles or nearly straddles zero for both
bases, and neither the Wilcoxon nor the sign test rejects at
$\alpha=0.05$. The frontier's rise (Table~\ref{tab:results-4}, Figure~\ref{fig:results-frontier}) should be read as a
descriptive pattern in the means across the grid, not as a difference this
design establishes is distinguishable from noise.

\subsection{The pair-level test: a non-inferential illustration of pseudoreplication}
\label{sub:results-the-pair-level-test-a-non-inferential-illustration-of-pseudoreplication}

This subsection is not evidence for or against any claim in this paper; we
include it only as a worked illustration of why the pairing above is
wrong, for a reader who wants to see the contrast concretely rather than
take our word for it. As a sensitivity analysis illustrating why naïve
independence assumptions can be misleading, we also report the
pre-specified comparison that pairs
by (dataset, regime, seed, selection mode), treating all $90$ pairs per
base as independent — the two selection modes are not independent
replicates of a cell, only two readings of the same trained run, so this
overstates the effective sample size roughly tenfold. Under that
assumption the test is significant for both bases after Holm correction:
for the Brier base, mean macro-F1 is $0.8136$ at supercritical against
$0.8167$ at published (Wilcoxon $W=18.0$, $p=0.017$, Holm-adjusted
$p=0.034$); for the logarithmic base, $0.8133$ against $0.8162$
($W=18.0$, $p=0.017$, Holm-adjusted $p=0.034$). Table~\ref{tab:results-3} reports this in
full. We give it specifically to show the contrast: the same underlying
differences, at the same mean magnitude, cross the conventional
significance threshold under the pair-level count and do not under the
cell-level one — the pre-specified analysis is not wrong about the data,
only about how many independent observations the data contain.

\begin{table}[htbp]
\centering
\caption{Non-inferential illustration of pseudoreplication: macro-F1 at supercritical versus published, $90$ paired observations per base (dataset $\times$ regime $\times$ seed $\times$ selection mode), mean $\pm$ standard deviation across all $90$ pairs, Wilcoxon signed-rank, Holm-corrected across the two bases. This table is not evidence; it exists to show what treating pseudoreplicated pairs as independent does to a $p$-value. The basis for this paper's conclusions is Table~\ref{tab:results-2}.}
\label{tab:results-3}
\begin{tabular}{lcccccc}
\hline
Base & supercritical & published & Paired diff. & $W$ & $p$ (raw) & $p$ (Holm) \\
\hline
Brier & $0.8136\pm0.1967$ & $0.8167\pm0.1955$ & $-0.0031\pm0.0114$ & $18.0$ & $0.017$ & $0.034$ \\
Logarithmic & $0.8133\pm0.1954$ & $0.8162\pm0.1952$ & $-0.0029\pm0.0105$ & $18.0$ & $0.017$ & $0.034$ \\
\hline
\end{tabular}
\end{table}

The standard deviation here is large relative to the mean difference
because it is taken across all $90$ pairs, which mix informative and
non-informative regimes at very different baseline macro-F1.

Taken alone, emnist\_balanced's own $10$ pairs (Brier base) do not show a
significant difference under the pair-level count either (mean supercritical
$0.8907$, mean published $0.8900$, $8$ of $10$ pairs exactly tied, Wilcoxon
$p=0.50$); at $c=47$, this design does not confirm the same
monotone-past-$\lambda^\ast$ pattern the $c=10$ regimes show — see below.

\subsection{Exploratory: macro-F1, ECE, and NLL across the full $\lambda$ grid}
\label{sub:results-exploratory-macro-f1-ece-and-nll-across-the-full-lambda-grid}

Everything in this subsection is exploratory, in a sense the primary
comparison above is not: the only point on the $\lambda$ grid this design's
formal test compares is supercritical against published (Table~\ref{tab:results-2}), at the
cell level. The full trajectory across every grid point is reported as
means with no formal test of the trajectory itself — no hierarchical or
trajectory-level model has been fit to it. We report the shape because it
is what the pre-specified design was built to look at, not because a
$p$-value attaches to the shape claim; readers should weigh it as a
descriptive pattern in aggregate means, not as a statistically established
characterisation of the frontier.

The estimand this subsection addresses is also conditional, not
unconditional: it pools only the regimes that clear the informativeness
threshold ("Which regimes inform the frontier," above), so it describes
the frontier's shape among regimes
difficult enough for $\lambda$ to have room to matter, not an average over
every regime this design ran. The four regimes it excludes are reported on
their own terms, separately, at the end of this subsection — their own
estimand is "what happens once accuracy has already saturated," a
different question with a different (and, there, flat) answer. Neither
estimand stands in for the other.

This subsection pools only the four $c=10$ informative regimes
(mnist/hard, cifar10/subset, cifar10/noisy, cifar10/hard): mnist, cifar10,
and emnist\_balanced use different $\lambda$ grids, since $\lambda^\ast$ and
every other grid point are $c$-dependent (Amendment 1), and pooling absolute
$\lambda$ values across different grids would silently average scores
measured at different penalties under the same grid-point name. We give
emnist\_balanced's own frontier separately, below, rather than blend it in
here.

\begin{figure}[htbp]
\centering
\includegraphics[width=3.4in]{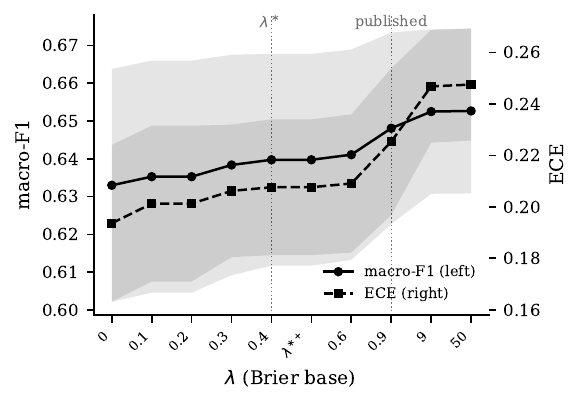}
\caption{Mean macro-F1 (left axis, solid) and mean ECE (right axis,
dashed), each with a shaded band of $\pm1$ seed-to-seed standard
deviation averaged across the four pooled regimes (the same dispersion
Table~\ref{tab:results-4} reports numerically), across the Brier-base $\lambda$ grid,
checkpointing selection, pooled over the four $c=10$ informative regimes
(mnist/hard, cifar10/subset, cifar10/noisy, cifar10/hard) —
$\lambda^\ast=0.400$ and the published penalty $0.900$ marked. Both
measures rise together across the entire grid; there is no point at which
macro-F1 continues improving while ECE holds steady. The bands are wide
relative to the rise itself, consistent with the text's own description
of this trajectory as descriptive rather than statistically established.
emnist\_balanced ($c=47$, a different $\lambda$ grid) is not pooled into
this figure — see the text for its own frontier. Data are derived from
the frozen experimental records for the cells trained under the original
pre-specified configuration. Unlike Figures 1–4, this figure is
experimental, not theoretical, and was built only after the pre-specified
sweep completed.}
\label{fig:results-frontier}
\end{figure}

Pooled across the four $c=10$ informative regimes, checkpointing selection,
mean macro-F1 rises across the entire $\lambda$ grid for both bases, and
ECE and NLL rise alongside it rather than holding steady past $\lambda^\ast$;
neither curve plateaus. Table~\ref{tab:results-4} gives every value; Figure~\ref{fig:results-frontier} plots the
Brier base's macro-F1 and ECE. Early-stopping selection reproduces the same
shape (Table~\ref{tab:results-5}), so it is not an artefact of the selection rule.

\begin{table}[htbp]
\centering
\caption{Macro-F1, ECE, and test NLL across the $\lambda$ grid, pooled over the four $c=10$ informative regimes, checkpointing selection. Dispersion is the seed-to-seed standard deviation at a fixed (dataset, regime, $\lambda$) cell, averaged across the four regimes — it reflects seed noise within a cell, not the much larger spread between regimes of very different difficulty. The implemented grid point has no logarithmic row: the mean/sum-form implementation gap is Brier-specific (relation to published rules).}
\label{tab:results-4}
\begin{tabular}{lccccc}
\hline
$\lambda$ & value & Base & Macro-F1 & ECE & NLL \\
\hline
zero & $0$ & Brier & $0.6330\pm0.0308$ & $0.194\pm0.031$ & $1.230\pm0.049$ \\
zero & $0$ & Log & $0.6263\pm0.0327$ & $0.176\pm0.032$ & $1.227\pm0.047$ \\
critical ($\lambda^\ast$) & $0.4$ & Brier & $0.6397\pm0.0280$ & $0.208\pm0.026$ & $1.235\pm0.049$ \\
critical ($\lambda^\ast$) & $1.609438$ & Log & $0.6409\pm0.0286$ & $0.209\pm0.027$ & $1.235\pm0.050$ \\
supercritical & $0.400001$ & Brier & $0.6397\pm0.0280$ & $0.208\pm0.026$ & $1.235\pm0.049$ \\
supercritical & $1.609439$ & Log & $0.6409\pm0.0286$ & $0.209\pm0.027$ & $1.235\pm0.050$ \\
published & $0.9$ & Brier & $0.6481\pm0.0253$ & $0.225\pm0.029$ & $1.246\pm0.058$ \\
published & $2.303$ & Log & $0.6482\pm0.0254$ & $0.226\pm0.028$ & $1.245\pm0.058$ \\
implemented & $9.0$ & Brier & $0.6525\pm0.0218$ & $0.247\pm0.022$ & $1.270\pm0.059$ \\
limit & $50.0$ & Brier & $0.6527\pm0.0219$ & $0.248\pm0.022$ & $1.271\pm0.059$ \\
limit & $50.0$ & Log & $0.6527\pm0.0219$ & $0.248\pm0.022$ & $1.271\pm0.059$ \\
\hline
\end{tabular}
\end{table}

\begin{table}[htbp]
\centering
\caption{Early-stopping selection, macro-F1 only, same four regimes: the shape survives a change of selection rule. The $\lambda$ column gives the Brier-base value; the log base uses its own value at each named grid point (Table~\ref{tab:experimental-design-1}), omitted here to keep the table compact.}
\label{tab:results-5}
\begin{tabular}{lccc}
\hline
$\lambda$ & value & Brier macro-F1 & Log macro-F1 \\
\hline
zero & $0$ & $0.6344$ & $0.6294$ \\
critical ($\lambda^\ast$) & $0.4$ & $0.6419$ & $0.6443$ \\
supercritical & $0.400001$ & $0.6419$ & $0.6443$ \\
published & $0.9$ & $0.6481$ & $0.6482$ \\
implemented & $9.0$ & $0.6525$ & n/a \\
limit & $50.0$ & $0.6527$ & $0.6527$ \\
\hline
\end{tabular}
\end{table}

Figure~\ref{fig:results-frontier} gives the same picture as Table~\ref{tab:results-4}: macro-F1
rises monotonically across the full grid, and ECE rises alongside it rather
than holding steady past $\lambda^\ast$.

Calibration and likelihood move in the same direction as $\lambda$ grows, not
the opposite one. Mean ECE (Brier base, $c=10$ informative regimes,
checkpointing) rises from $0.194$ at $\lambda=0$ to $0.208$ at supercritical,
$0.225$ at published, and $0.248$ at the limit; mean test NLL rises from
$1.230$ to $1.235$, $1.246$, and $1.271$ over the same points. The
logarithmic base shows the same monotone rise in both measures. ECE is one
specific, binning-dependent calibration diagnostic (Section~\ref{sub:experimental-design-outcome-measures} gives its exact construction), not calibration in the abstract, and we do
not have a second, differently-constructed calibration measure in this
design; the degradation claim is stronger for being corroborated by NLL,
a genuinely different quantity — a strictly proper scoring rule with no
binning choice of its own — showing the same monotone direction, but we
still name the specific measure rather than claim calibration in general
degrades by every possible construction of it. There is no
$\lambda$ range in this design where macro-F1 improves while calibration
holds steady, by either measure: every increment in $\lambda$ that buys
macro-F1 costs
calibration, throughout the grid, not only beyond some threshold.

\subsection{The frontier at a different class count ($c=47$): a case study}
\label{sub:results-the-frontier-at-a-different-class-count-c-47-a-case-study}

emnist\_balanced was added by Amendment 1, after the original two-dataset
design was pre-specified, specifically to test whether this pattern holds
at a class count other than $c=10$. We present this as an exploratory case
study, one dataset at one additional class count, rather than a
confirmatory test of a class-count effect — it was not part of the design
frozen before the first result was read, and a single dataset cannot by
itself establish one. It is also not a like-for-like comparison population:
four of the five informative $c=10$ cells are reduced-data or noisy
regimes, constructed to be difficult, while this is the one informative
cell at full training strength on its dataset's natural difficulty (see
"Which regimes inform the frontier," above). The two class counts differ
in how the
classification task was made hard, not only in $c$, so a disagreement
between them is at least as readily read as a construction difference as a
class-count effect. We therefore treat this cell as a deliberately
exploratory counterexample to universality — it shows the $c=10$ pattern
does not hold everywhere, without identifying why — rather than as
evidence of a $c$-dependent phenomenon specifically. Its own grid cannot
be pooled with the above (different $\lambda$ values), so we report it as
its own, separate frontier in Table~\ref{tab:results-6}: flat, not rising, through
supercritical and subcritical, before macro-F1 falls slightly and ECE and
NLL both rise from published onward. Calibration still degrades
monotonically here, matching the $c=10$ regimes, but macro-F1 does not
keep improving to compensate — if anything it very slightly worsens past
published, the opposite direction from the $c=10$ finding, though within
noise given the small sample (paired comparison above: $p=0.50$, not
significant).

\begin{table}[htbp]
\centering
\caption{emnist\_balanced/full ($c=47$), Brier base, checkpointing — its own $\lambda$ grid, not comparable in absolute value to Tables~\ref{tab:results-4}--\ref{tab:results-5}. Rows from zero through subcritical are identical (the same checkpoint is selected throughout), so only the distinct values are shown.}
\label{tab:results-6}
\begin{tabularx}{\textwidth}{Y>{\centering\arraybackslash}p{1.02in}>{\centering\arraybackslash}p{1.17in}>{\centering\arraybackslash}p{1.02in}>{\centering\arraybackslash}p{1.02in}}
\hline
$\lambda$ & value & Macro-F1 & ECE & NLL \\
\hline
zero through subcritical & $0$ to $0.718$ & $0.8907\pm0.0022$ & $0.014\pm0.004$ & $0.309\pm0.005$ \\
published & $0.979$ & $0.8900\pm0.0029$ & $0.015\pm0.005$ & $0.314\pm0.013$ \\
implemented and limit & $46.0$, $200.0$ & $0.8898\pm0.0030$ & $0.019\pm0.006$ & $0.318\pm0.017$ \\
\hline
\end{tabularx}
\end{table}

We do not read this as contradicting the $c=10$ result, and we do not paper
over the difference either: at $c=47$, in this one regime, the honest
reading is that increasing $\lambda$ past $\lambda^\ast$ buys no measured
macro-F1 gain here, only a calibration cost — closer to the pattern
flat predicted than to rising. Whether this reflects something
genuinely different about how class count interacts with the frontier's
shape, or is a property of this one dataset and regime rather than of $c=47$
generally, cannot be distinguished from a single additional dataset.

On the four excluded, non-informative regimes, macro-F1 is flat to within
three parts in a thousand across the entire grid ($0.9678$ at $\lambda=0$ to
$0.9680$ at the limit, Brier base, checkpointing) — the pattern the design
anticipated for regimes where macro-F1 has already saturated, and a check
that the informativeness threshold is doing the right thing rather than an
arbitrary cut.

\subsection{Baselines: does the frontier require superiority, or only re-weighting?}
\label{sub:results-baselines-does-the-frontier-require-superiority-or-only-re-weighting}

The weighted mixture originally intended as a no-superiority control is
\begin{equation}M_w(p,\omega) = (1-w)\cdot\bar S_{BS}(p,\omega) + w\cdot\mathrm{err}(p,\omega),\end{equation}
where $\bar S_{BS}$ is mean-form Brier, $\mathrm{err}$ is the $0$–$1$
misclassification indicator, and $w\in\{0, 0.25, 0.5, 0.75, 0.9, 1.0\}$.
Under Convention B, $\mathrm{err}(p,\omega)=\mathbb{1}[p\in\xi]$ exactly —
the same indicator the family's own penalty term uses — so, writing
$\bar S_{BS}=S_{BS}/c$ (mean/sum-form, preliminaries) and collecting terms,
\begin{equation}M_w(p,\omega) = \frac{1-w}{c}\Big[S_{BS}(p,\omega) + \frac{wc}{1-w}\,\mathbb{1}[p\in\xi]\Big]
= \frac{1-w}{c}\,S_{\lambda(w)}(p,\omega),\end{equation}
\begin{equation}\lambda(w):=\frac{wc}{1-w}.\end{equation}
For every $w\in[0,1)$, $M_w$ is therefore a strictly positive scalar
multiple of a member of the very family $S_\lambda$ it was meant to test
against — not an independent construction. This has two consequences we
did not anticipate when the design was pre-specified: $M_w$ ranks any fixed
set of checkpoints identically to $S_{\lambda(w)}$ (a positive scalar
does not change an argmin), and $M_w$ is itself superior and strictly
proper whenever $\lambda(w)>\lambda^\ast$, since both properties are
preserved under positive scaling. Solving $\lambda(w)>\lambda^\ast$ gives
$w>w^\ast:=\lambda^\ast/(\lambda^\ast+c)$; at $c=10$, $\lambda^\ast=0.4$, this is
$w^\ast\approx0.038$ (Table~\ref{tab:results-7}), so every weight actually tested here except
$w=0$ is already superior. We verified the identity numerically to
floating-point precision and confirm it is exact, not approximate.

\begin{table}[htbp]
\centering
\caption{The weight-to-$\lambda$ map $\lambda(w)=wc/(1-w)$ at $c=10$, for every weight tested, against the superiority threshold $w^\ast\approx0.038$ ($\lambda^\ast=0.4$).}
\label{tab:results-7}
\begin{tabular}{lcc}
\hline
$w$ & $\lambda(w)$ & Superior ($\lambda(w)>\lambda^\ast$)? \\
\hline
$0$ & $0$ & no \\
$0.25$ & $3.33$ & yes \\
$0.5$ & $10.0$ & yes \\
$0.75$ & $30.0$ & yes \\
$0.9$ & $90.0$ & yes \\
$1.0$ & $\infty$ & yes (limit) \\
\hline
\end{tabular}
\end{table}

This means $M_w$ does not do what we designed it to do. Its checkpoint
selections agreeing closely with $S_\lambda$'s own — which we reported
last as an empirical finding — is a necessary consequence of this identity,
not independent confirmation (Table~\ref{tab:results-8}): at $w=0$ versus $\lambda=0$ (Brier
base, the same rule under either name up to the constant), the two select
the identical checkpoint in all $90$ (dataset, regime, seed, mode)
combinations. At $w=1$ versus the $\lambda\to\infty$ limit, agreement is
still high but no longer exact, the small remainder attributable to
$\lambda=50$ approximating rather than reaching the limit exactly and, for
the logarithmic case, to comparing $M_w$ (still Brier-built) against a
genuinely different base rule. Because $M_w$'s own superiority is a property of
$M_w$ itself, not of which family it is being read against, this also
removes it as a no-superiority control for the logarithmic-base
comparison: it is superior in the same absolute sense there too, even
though its checkpoint selections are not an algebraic identity against the
log family's own $S_\lambda$. We do not, therefore, read the frontier
below as isolating superiority from error-oriented re-weighting, and we
withdraw that framing. The honest description is narrower — the observed
benefit is consistent with increasing weight on misclassification, and
this design does not provide independent evidence that superiority is
unnecessary for it, because the instrument built to supply that evidence
turned out, on inspection, to already carry the property it was meant to
be free of.

\begin{table}[htbp]
\centering
\caption{Checkpoint agreement between $M_w$ and the corresponding member of $S_\lambda$, at the two endpoints where the identity above applies exactly ($w=0$) or approximately ($w=1$), out of $90$ (dataset, regime, seed, mode) combinations.}
\label{tab:results-8}
\begin{tabular}{lcc}
\hline
Comparison & Base & Agreement \\
\hline
$w=0$ vs. $\lambda=0$ & Brier & $90/90$ ($100\%$) \\
$w=1$ vs. $\lambda\to\infty$ (limit) & Brier & $82/90$ ($91\%$) \\
$w=1$ vs. $\lambda\to\infty$ (limit) & Logarithmic & $81/90$ ($90\%$) \\
\hline
\end{tabular}
\end{table}

Reported only as what was measured, without the mechanism-isolation
reading above: unlike the frontier tables above, baselines are indexed by
weight, not by $\lambda$, so pooling across all five informative regimes —
the four $c=10$ regimes and emnist\_balanced — raises no grid-mismatch
concern. Mean macro-F1 (checkpointing) is $0.6845$ at weight $0$, rising to
$0.6992$ at $0.25$, $0.7000$ at $0.5$, and $0.7001$ at weights $0.75$ and
above — the same curve $S_\lambda$ itself traces at the corresponding
$\lambda(w)$, by the identity above, not an independently arrived-at
match. Monitoring error rate directly gives $0.7001$, and monitoring
macro-F1 directly gives $0.7006$, close to the same ceiling $S_\lambda$
reaches as $\lambda\to\infty$; these two are separate criteria, not
special cases of $M_w$, and remain informative as endpoints even though
$M_w$ itself no longer serves as a mechanism-isolating control. Table~\ref{tab:results-9}
gives every value.

\begin{table}[htbp]
\centering
\caption{Baseline mean macro-F1, all five informative regimes ($c=10$ and $c=47$ together — weight, unlike $\lambda$, is on the same scale for every dataset), checkpointing. "$M_w$" labels the weighted mixture; despite its original design intent, it is not a no-superiority rule except at $w=0$ (baselines, above).}
\label{tab:results-9}
\begin{tabularx}{\textwidth}{Y>{\centering\arraybackslash}p{0.76in}>{\centering\arraybackslash}p{0.93in}}
\hline
Baseline & Weight & Macro-F1 \\
\hline
monitor macro-F1 directly & — & $0.7006$ \\
monitor error rate directly & — & $0.7001$ \\
$M_w$ & $0.00$ & $0.6845$ \\
$M_w$ & $0.25$ & $0.6992$ \\
$M_w$ & $0.50$ & $0.7000$ \\
$M_w$ & $0.75$ & $0.7001$ \\
$M_w$ & $0.90$ & $0.7001$ \\
$M_w$ & $1.00$ & $0.7001$ \\
\hline
\end{tabularx}
\end{table}

\subsection{Stability: how consistently each $\lambda$ selects the same checkpoint}
\label{sub:results-stability-how-consistently-each-lambda-selects-the-same-checkpoint}

The seed-to-seed standard deviation of the selected epoch (Table~\ref{tab:results-10}) shows
no consistent direction with $\lambda$: it falls for mnist/hard and
cifar10/hard, rises then falls for cifar10/noisy, and is essentially flat
for cifar10/subset and emnist\_balanced/full. We do not read a trend into
this: the pattern differs by regime and does not move monotonically with
$\lambda$ the way the four outcome measures above do.

\begin{table}[htbp]
\centering
\caption{Seed-to-seed standard deviation of the selected epoch, Brier base, checkpointing, at three grid points per regime.}
\label{tab:results-10}
\begin{tabular}{lccc}
\hline
Regime & $\lambda=0$ & supercritical & published \\
\hline
mnist/hard & $6.67$ & $6.28$ & $6.19$ \\
cifar10/hard & $3.78$ & $2.07$ & $2.07$ \\
cifar10/noisy & $3.05$ & $3.19$ & $1.34$ \\
cifar10/subset & $1.10$ & $1.10$ & $1.10$ \\
emnist\_balanced/full & $0.89$ & $0.89$ & $1.14$ \\
\hline
\end{tabular}
\end{table}

\subsection{Summary: which pre-specified outcome obtained}
\label{sub:results-summary-which-pre-specified-outcome-obtained}

Of the three outcomes stated as equally acceptable in advance, the \textit{means}
at $c=10$ match rising descriptively: selected-model macro-F1 increases
across the whole $\lambda$ grid, including past the published value, rather
than flattening at $\lambda\ge\lambda^\ast$ (flat) or peaking at an interior
point (peaked). We do not, however, treat this as a statistically
established result: the cell-level analysis above shows this pattern is not
distinguishable from noise once the seed/mode nesting is accounted for, so
what obtained is a descriptive match to rising's shape, not a
confirmed rejection of flat in the statistical sense the pre-specified
test alone would have suggested. With that caveat, the pattern is still a
negative result for the paper's own motivating hypothesis, stated in the
same register it would carry if favourable: the published penalty shows a
higher mean macro-F1 than a smaller one, descriptively, in a design that
no longer has a valid instrument for asking why (baselines, above). What the data do show,
independent of that open question, is that monitoring error rate or
macro-F1 directly reaches the same ceiling without the calibration cost
that grows alongside $\lambda$ — a fact about where the ceiling sits, not
evidence for either mechanism.

At $c=47$, an exploratory analysis added by Amendment 1 specifically to
test whether this generalises, the same design instead produced a pattern
closer to flat:
macro-F1 essentially unchanged past $\lambda^\ast$, calibration still
degrading. We state both outcomes rather than let the larger, more
data-rich $c=10$ result stand in for a class-count question it was not
designed to answer alone: with one dataset at $c=47$ against four regimes
at $c=10$, this design cannot distinguish whether the qualitative pattern
genuinely depends on class count or on which dataset and regime happen to
be tested. We return to what both outcomes do and do not imply for the
source paper's own contribution in the discussion that follows.


\section{Discussion}
\label{sec:discussion}

\subsection{What the empirical outcome means for practitioners}
\label{sub:discussion-what-the-empirical-outcome-means-for-practitioners}

The results above are a two-dataset case study, not a population-level
experiment, and we read them that way here. The frontier observed at
$c=10$ matches rising, not the flat outcome we had pre-specified as our
own motivating hypothesis (experimental design): mean selected-model
macro-F1 rises as $\lambda$ grows past $\lambda^\ast$, though not
distinguishably from noise once seed/selection-mode nesting is accounted
for (Section~\ref{sub:results-primary-comparison-supercritical-vs-published-macro-f1-cell-level-analysis}), and a fifth regime at $c=47$ does not
reproduce even this descriptive pattern. Two things bound what this buys
a practitioner. First, the theory establishes admissibility — which
$\lambda$ carries the guarantee — not optimality under macro-F1 or any
other downstream criterion. Second, the weighted-mixture baseline meant
to isolate superiority's own contribution from re-weighting toward
misclassification does not (Section~\ref{sub:results-baselines-does-the-frontier-require-superiority-or-only-re-weighting}): it turned out to be a
positive rescaling of a member of the family under test, not the
genuinely non-equivalent counterfactual isolating that contribution would
need, so neither the $c=10$ pattern nor its absence at $c=47$ can be
attributed to superiority specifically. We do not read the guarantee as
diminished by this — a metric that can rank a wrong answer ahead of a
right one is defective regardless of what correlates with fixing it —
only as less mechanistically understood than our design assumed. What
would make it legible is a genuinely independent no-superiority
construction, and a decision-theoretic property linking the ordering
guarantee to a specific downstream loss — for instance, that a superior
metric's checkpoint selection is risk-dominant under a stated class of
misclassification costs; we leave both as directions this paper does not
resolve, not as experiments already carried out.

\subsection{The pointwise guarantee versus finite-sample selection}
\label{sub:discussion-the-pointwise-guarantee-versus-finite-sample-selection}

Theorems~\ref{thm:strict-propriety} and~\ref{thm:minimality} (Section~\ref{sec:theory-2}) are pointwise guarantees, stated here
under the single-valued tie convention (Convention B) this paper
recommends and this experiment uses throughout — Theorem~\ref{thm:strict-propriety}'s strict
propriety is specific to that convention, since Convention A and
Convention C each fail it (Propositions~\ref{prop:convention-a-failure} and~\ref{prop:convention-c-failure}, theory part 2): $S_\lambda$
is strictly proper for every $\lambda\ge0$ (Theorem~\ref{thm:strict-propriety}), and superior for
every $\lambda>\lambda^\ast$ (Theorem~\ref{thm:minimality}), regardless of how small
$\lambda-\lambda^\ast$ is. Model selection, by contrast, compares
checkpoints by their average score over a finite validation set of $N$
examples, and it is this finite-sample comparison, not the pointwise
guarantee by itself, that determines what a practitioner's pipeline actually
does. Writing $k$ for the number of validation examples a given checkpoint
misclassifies under the single-valued tie convention, the family's own
definition ($S_\lambda=S_{\text{base}}+\lambda\cdot\mathbb{1}[p\in\xi]$,
theory part 2) gives its mean validation score as
$\bar S_{\text{base}}+\lambda\cdot k/N$. Comparing two checkpoints $A,B$, the
penalty term contributes $\lambda\cdot(k_A-k_B)/N$ to their score gap — an
integer multiple of $\lambda/N$, since $k_A,k_B$ are non-negative integers.
A tiny $\epsilon=\lambda-\lambda^\ast$ therefore perturbs the finite-sample
comparison by at most a term of order $\epsilon/N$: enough to satisfy the
pointwise guarantee at every $\lambda>\lambda^\ast$, but not necessarily
enough to change which checkpoint a given validation set selects, since
that additionally requires $\epsilon/N$ to exceed whatever gap already
separates the near-tied checkpoints on the base score alone. Table~\ref{tab:results-4}'s
critical and supercritical rows (Section~\ref{sec:results}) illustrate this directly: at
$\epsilon=10^{-6}$, the two rows agree to four decimal places across every
regime and both bases, because $\epsilon/N$ is far smaller than any
base-score gap in this design's validation sets. Nothing about this is a
failure of the guarantee — $S_{\lambda^\ast+\epsilon}$ genuinely is superior
for every $\epsilon>0$ — it is a statement about how small a finite-sample
effect can be while the pointwise property it certifies still holds
exactly. Choosing $\epsilon$ larger, as the published penalty explores, is
a separate, empirical question about how much of a finite-sample effect a
practitioner wants, not a further requirement of the guarantee itself.

\subsection{Recommendations}
\label{sub:discussion-recommendations}

$\lambda>\lambda^\ast$ defines a region where the superiority and
strict-propriety guarantees hold, not a region of equivalent practical
behaviour: this design's own frontier (Section~\ref{sec:results}) shows ECE and NLL
degrading measurably, and checkpoint selection changing, across that
entire region. We recommend any $\lambda>\lambda^\ast$ as the default
whenever those two guarantees alone are required, on the theoretical
grounds established earlier in the paper: every such $\lambda$ carries
them under the recommended tie convention, and the admissible open
interval $(\lambda^\ast,\infty)$ has no smallest element (Theorem~\ref{thm:minimality}, theory
part 2), so the argument does not single out one $\lambda$ as minimal —
our own grid instantiates it with supercritical, $\lambda^\ast+\epsilon$ for
$\epsilon=10^{-6}$, purely as a computational demonstration that the
strict inequality can hold without materially changing finite-sample
checkpoint selection (previous subsection), not as a further
recommendation in its own right. A practitioner adopting this default
should pick any convenient $\lambda$ strictly greater than $\lambda^\ast$;
nothing in the guarantee improves by choosing $\epsilon$ smaller. This
recommendation is theoretical: it holds regardless of class count or
dataset, and different admissible $\lambda$ are not thereby practically
equivalent — different values still produce measurable calibration,
likelihood, and checkpoint-selection differences, as the frontier shows.

We adopt an order-induced tie-break specifically (Convention B,
preliminaries), rather than randomising each tie or treating it as
negligible, because it is deterministic and reproducible from a stated
order alone, and matches standard argmax practice in classification
software; Corollary~\ref{cor:order-independence} (Section~\ref{sec:theory-2}) shows the boundary result does not
depend on \textit{which} order is chosen, only on there being one. Randomised
tie-breaking would need Lemma~\ref{lem:penalty-proper}'s propriety argument extended to an
expected penalty under the randomisation, which we have not carried out;
treating a tie as negligible is Convention A's own position, and
Proposition~\ref{prop:convention-a-failure} (Section~\ref{sec:theory-2}) shows that position fails strict propriety
once $\lambda$ exceeds a $Q$-dependent threshold. We record the
comparison to randomised tie-breaking as an open direction, not a settled
one.

The condition under which to depart from this guarantee-preserving
default is exactly the one our $c=10$ data point to, descriptively, not
one they establish: if the objective is macro-F1-driven selection
specifically, our $c=10$ frontier shows further gains in the observed
means by increasing $\lambda$ well past supercritical, up to the point of
monitoring error rate directly — though these are means, not a
statistically confirmed effect, and this design has no instrument that
attributes them to superiority rather than to the penalty's re-weighting
generally (previous subsection) — while our $c=47$ frontier shows no such
gain in the one regime tested there, so this part of the recommendation
is conditional on the $c=10$ pattern replicating, not established
generally. Nothing in the theory distinguishes the published value's
particular magnitude as a stopping point along either path — it was set
from $\Sigma_\psi$ alone (Corollary~\ref{cor:positioning-published}, theory, part 2), not from where any
measured frontier plateaus. Where the $c=10$ pattern holds, a practitioner
willing to act on it despite these caveats has an observed route to
macro-F1 gains — monitoring error rate directly, accepting the
calibration cost explicitly — while where it does not, as at $c=47$,
supercritical already sits at or near whatever macro-F1 ceiling is
available in the data, with no observed benefit to moving past it.

\subsection{Directions not pursued here}
\label{sub:discussion-directions-not-pursued-here}

Two further directions remain open. First, $S_\lambda$ generalises to a
full misclassification cost matrix in place of the scalar penalty —
including the special case of one scalar $\lambda_\omega$ per true class,
which would let a class-dependent penalty counteract macro-F1's own
per-class averaging — with propriety plausibly preserved by the same
argument that establishes it for the scalar case; we state this only as a
direction, since we have not carried out that argument, and this design
was not built to test it. Second, the source paper's own Future Works
section identifies multi-label classification, hierarchical
classification, calibration research, and ordinal classification as
directions; our contribution is orthogonal to these — we do not extend
the domain the source paper's scoring rules apply to, and instead treat
the penalty magnitude, within the single-label multi-class setting the
source paper already addresses, as a parameter that had not been
calibrated against the rule's own range.


\section{Limitations}
\label{sec:limitations}

\subsection{The no-superiority baseline is not independent of the family it was meant to test against}
\label{sub:limitations-the-no-superiority-baseline-is-not-independent-of-the-family-it-was-meant-to-test-against}

The weighted mixture $M_w$ (Section~\ref{sub:results-baselines-does-the-frontier-require-superiority-or-only-re-weighting}) was pre-specified as a
control lacking the superiority property, meant to isolate whether the
observed frontier is driven by that property or simply by re-weighting
toward misclassification. It does not do this. Under Convention B and
sum-form Brier, $M_w$ is exactly a positive scalar multiple of $S_\lambda$
at $\lambda(w)=wc/(1-w)$, so it inherits $S_\lambda$'s own checkpoint
rankings and its superiority whenever $\lambda(w)>\lambda^\ast$ — true for
every weight tested here except $w=0$. This design does not, in fact,
contain a valid no-superiority control; framing the frontier's rise as
"consistent with error-oriented re-weighting rather than superiority"
overstates what the baseline evidence supports, and we correct that
framing throughout (Section~\ref{sec:results}; Section~\ref{sec:discussion}). A genuinely independent
no-superiority construction — one not built from the same
misclassification indicator the family's own penalty uses — remains to be
designed; we do not have one.

\subsection{Tie-convention costs}
\label{sub:limitations-tie-convention-costs}

$S_\lambda$ is discontinuous wherever the argmax of the forecast changes:
a single validation sample crossing from one predicted class to another
can flip the penalty term for every forecast on that side of the
boundary. What this implies for the variance of a selected epoch is not
established here — we have not proved a bound relating discontinuity
size to how often a monitoring criterion changes which epoch it prefers.
Our own measurement of this variance (Section~\ref{sub:results-stability-how-consistently-each-lambda-selects-the-same-checkpoint}) is
inconclusive: the seed-to-seed standard deviation of the selected epoch
moves in different directions across regimes as $\lambda$ increases, with
no pattern consistent enough to confirm or rule out an effect.

A related cost: $\lambda^\ast$ is convention-independent as a value, but
whether it is itself admissible, and whether the resulting penalty is
strictly proper at every $Q$, are not simultaneously available under any
single tie-breaking rule. Under the permissive convention, $\lambda=\lambda^\ast$
is admissible but strict propriety fails once $\lambda$ exceeds a
$Q$-dependent threshold; under the ties-to-$\xi$ convention, strict
propriety fails outright for every $\lambda>0$ at every $Q$ with a tied
mode; under the recommended single-valued convention, strict propriety
holds for every $Q$, but $\lambda^\ast$ itself is not admissible, for any
choice of order. A reader who wants both properties from a single, fixed
$\lambda=\lambda^\ast$ under a single convention cannot have them; a choice
among these costs is unavoidable, not merely underspecified in this
paper.


\section{Conclusion}
\label{sec:conclusion}

We characterised superiority as separation between two closed-form
endpoints — the largest score a rule attains on a correctly classified
forecast and the smallest it attains on an incorrectly classified one —
for both the Brier and logarithmic scores. Their difference is the
infimal penalty $\lambda^\ast$: closed-form, proved, and the greatest lower
bound a penalty must clear for superiority to hold. The published penalty
is one specific point strictly above $\lambda^\ast$ within an unbounded
admissible range — sufficient for superiority, but not the smallest
penalty that is. The same characterisation yields a one-parameter family
that is strictly proper throughout under a single-valued tie convention,
repairing the published propriety argument for the whole family rather
than the two published instances.

Empirically, mean selected-model macro-F1 rose with the penalty past
$\lambda^\ast$ at ten classes, but not distinguishably from noise once
seed/selection-mode nesting is accounted for, and a fifth regime at
forty-seven classes did not reproduce even this descriptive pattern. The
baseline meant to test whether superiority itself, rather than the
penalty's re-weighting toward misclassification generally, drives this
pattern turned out to be algebraically inseparable from the family under
test. The experiment therefore does not establish that a penalty larger
than $\lambda^\ast$ improves model selection, and it does not establish a
causal role for superiority in whatever pattern it does show; both
questions remain open.

We recommend any $\lambda$ strictly above $\lambda^\ast$ whenever strict
propriety and superiority are the objective, and treat the choice among
admissible values for other purposes — accuracy, calibration, or any
other downstream criterion — as a separate, unresolved problem this
paper's theory does not settle.

\section*{Data and code availability}
The implementation of the closed-form quantities characterised in this
paper, the pre-specified experimental configuration and training pipeline,
the cached per-epoch validation predictions and the analysis script used to
produce every number reported here, random seeds, and the exact dataset
construction (train/validation/test splits, label-noise injection, and
reduced-data subsetting) are made available at
\url{https://github.com/Ruhallah93/superior-scoring-rules}.

\bibliography{refs}
\bibliographystyle{plainnat}

\end{document}